\documentclass[letterpaper]{article} 
\usepackage{aaai23} 
\usepackage{times}  
\usepackage{helvet}  
\usepackage{courier}  
\usepackage[hyphens]{url}  
\usepackage{graphicx} 
\usepackage{natbib}  
\usepackage{caption} 
\usepackage{algorithm}
\usepackage{algorithmic}
\usepackage{amsfonts}
\usepackage{amsmath}
\usepackage{threeparttable}
\usepackage{booktabs}
\usepackage{subfig}
\usepackage{placeins}

\usepackage{newfloat}
\usepackage{listings}
\DeclareCaptionStyle{ruled}{labelfont=normalfont,labelsep=colon,strut=off} 
\floatstyle{ruled}
\newfloat{listing}{tb}{lst}{}
\floatname{listing}{Listing}
\title{AIO-P: Expanding Neural Performance Predictors\\ Beyond Image Classification}

\author{
    Keith G. Mills\textsuperscript{\rm 1,2}\thanks{Work done during an internship at Huawei.}, Di Niu\textsuperscript{\rm 1}, Mohammad Salameh\textsuperscript{\rm 2}, Weichen Qiu\textsuperscript{\rm 1}, Fred X. Han\textsuperscript{\rm 2}, \\Puyuan Liu\textsuperscript{\rm 2}, Jialin Zhang\textsuperscript{\rm 3}, Wei Lu\textsuperscript{\rm 2}, Shangling Jui\textsuperscript{\rm 3}
}
\affiliations{
    \textsuperscript{\rm 1}Department of Electrical and Computer Engineering, University of Alberta\\
    \textsuperscript{\rm 2}Huawei Technologies, Edmonton, Alberta, Canada\\
    \textsuperscript{\rm 3}Huawei Kirin Solution, Shanghai, China\\
    \{kgmills, dniu, wqiu1\}@ualberta.ca\\
    \{mohammad.salameh, fred.xuefei.han1, puyuan.liu, jui.shangling\}@huawei.com\\
    \{zhangjialin10, robin.luwei\}@hisilicon.com\\
}

\begin{document}

\maketitle

\begin{abstract}
Evaluating neural network performance is critical to deep neural network design but a costly procedure. Neural predictors provide an efficient solution by treating architectures as samples and learning to estimate their performance on a given task. However, existing predictors are task-dependent, predominantly estimating neural network performance on image classification benchmarks. They are also search-space dependent; each predictor is designed to make predictions for a specific architecture search space with predefined topologies and set of operations. In this paper, we propose a novel All-in-One Predictor (AIO-P), which aims to pretrain neural predictors on architecture examples from multiple, separate computer vision (CV) task domains and multiple architecture spaces, and then transfer to unseen downstream CV tasks or neural architectures. We describe our proposed techniques for general graph representation, efficient predictor pretraining and knowledge infusion techniques, as well as methods to transfer to downstream tasks/spaces. Extensive experimental results show that AIO-P can achieve Mean Absolute Error (MAE) and Spearman’s Rank Correlation (SRCC) below 1\% and above 0.5, respectively, on a breadth of target downstream CV tasks with or without fine-tuning, outperforming a number of baselines. Moreover, AIO-P can directly transfer to new architectures not seen during training, accurately rank them and serve as an effective performance estimator when paired with an algorithm designed to preserve performance while reducing FLOPs.
\end{abstract}

\section{Introduction}
\label{sec:intro}

Performance evaluation of neural network models is resource and time consuming. Several factors contribute to its expensiveness, such as task complexity, training dataset size, architecture topology, and training time. 
It is yet a main component and the bottleneck in Neural Architecture Search (NAS)~\cite{elsken2019neural}. Early NAS approaches train sampled architectures to completion during search \cite{zoph2017NAS} while later approaches adopt weight-sharing supernet approaches \cite{pham2018ENAS, liu2018DARTS, rezaei2021generative, mills2021l2nas, mills2021exploring}, which reduce the computational burden but do not eliminate it. Advanced supernet schemes like Once-for-All (OFA)~\cite{cai2020once} and BootstrapNAS~\cite{munoz2022automated} introduce progressive shrinking to train a reusable supernet. Specifically, OFA supernets are robust enough that individual architectures can be sampled for immediate evaluation on ImageNet~\cite{russakovsky2015imagenet}.

Zero-Cost Proxies (ZCP)~\cite{abdelfattah2021zero} are a recent development aiming to correlate performance with gradient statistics and thus can generalize to any type of network. However, the efficacy of ZCP methods depend on the architecture and task and may not be always reliable. 
Other recent schemes like NAS-Bench-301~\cite{zela2022surrogate}, SemiNAS~\cite{luo2020semi}, TNASP~\cite{lu2021TNASP} and WeakNAS~\cite{wu2021stronger} develop neural predictors that estimate architecture performance from network topology and operation features. However, they model customized architectures in specific search spaces, e.g., NAS-Bench-101~\cite{ying2019nasbench101} and 201~\cite{dong2020nasbench201} for common benchmark tasks like CIFAR image classification \cite{Krizhevsky09CIFAR}, and cannot be directly transferred to other challenging tasks like pose estimation or segmentation or to architectures with new types of topologies/connections.

\begin{figure}
	\centering
	\includegraphics[width=3.25in]{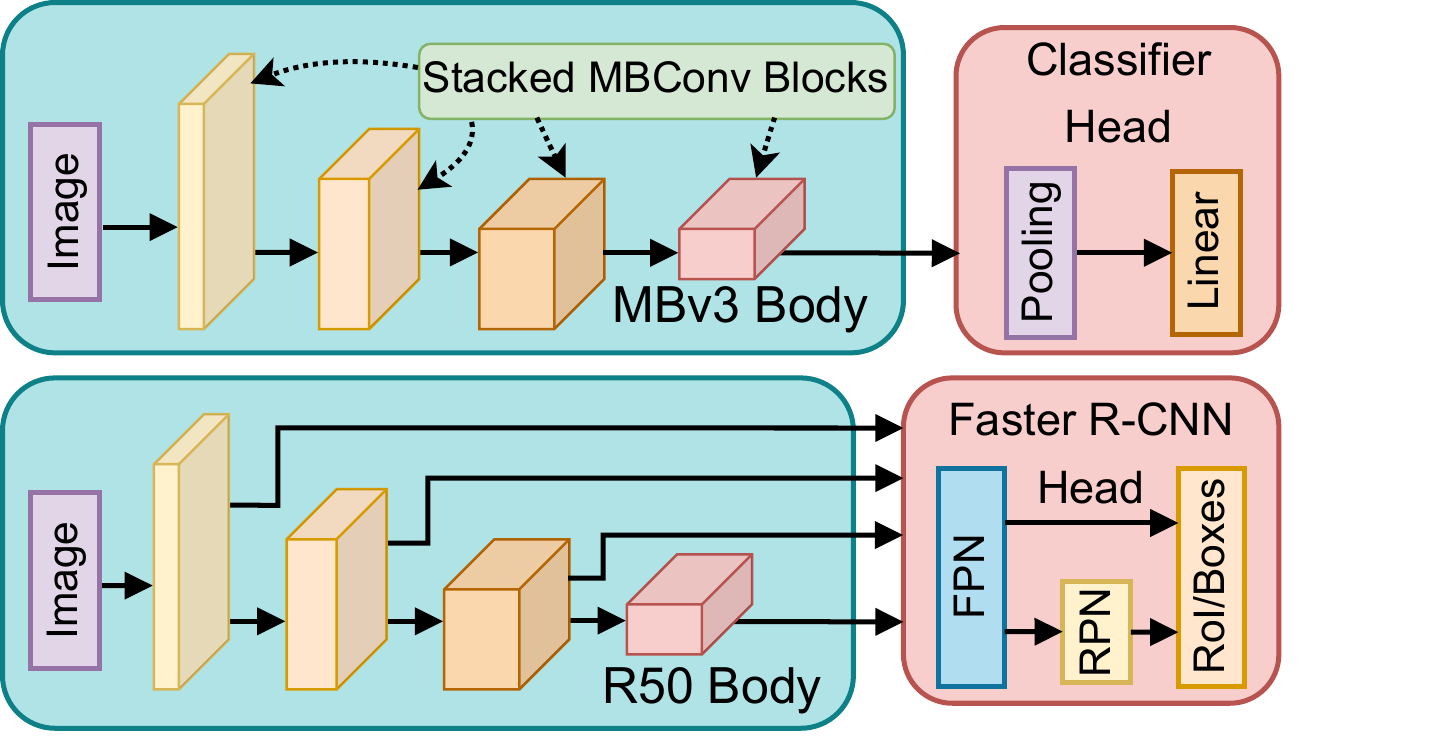}
	\caption{Network bodies from classification search spaces can pair with heads for different tasks. 
	A simple classification head has pooling and a linear layer. Faster R-CNN~\cite{ren2015faster} uses a Feature Pyramid Network (FPN) with 
	feature maps of different sizes. FPN 
	feeds a Regional Proposal Network (RPN) and a Region of Interest module (RoI) to 
	estimate bounding boxes.}  
	\label{fig:head_and_body}
\end{figure}

In this paper, we propose All-in-One Predictor (AIO-P), a multi-task neural performance predictor which achieves cross-task and cross-search-space transferability via predictor pretraining and domain-specific knowledge injection. AIO-P uses Computational Graphs (CG) to represent neural architectures, which is lower-level information extracted from TensorFlow execution and thus can model general types of architectures. We make the observation that many CV architectures consist of a \textit{body} (e.g., ResNet) that performs feature extraction and a \textit{head} that uses extracted features to generate task-specific outputs. Figure~\ref{fig:head_and_body} illustrates how architectures can be constructed for different tasks by combining various types of bodies and heads. Based on such network representations, we introduce an effective transfer learning scheme to first pretrain AIO-P on image classification benchmarks, then infuse domain knowledge from other tasks or architecture spaces efficiently, and finally transfer to a downstream task with minimum or no fine-tuning. Specifically, we propose the following techniques to achieve transferability to downstream tasks: 

First, we introduce $K$-Adapters~\cite{wang2021kadapt}, originally used to inject domain knowledge into language models, into a GNN predictor pretrained on network benchmarks for image classification (IC) such that the predictor can infuse knowledge from segmentation/detection tasks or other network topologies. We then transfer the learned model to perform predictions on potentially unseen downstream tasks or architectures. 

Second, we propose an efficient learning scheme to train AIO-P, and especially adapters, by a pseudo-labeling scheme for task-specific network performance. In order to reduce the high cost associated with labeling each individual architecture's performance on a given task, we propose to train a weight-sharing task head for all body architectures in an entire search space, e.g., all variants of MobileNetV3~\cite{howard2019searching} in OFA. After training the shared head, it can pair with any architecture body and fine-tune for a few minutes to obtain a pseudo-label that is positively correlated with the true label seen when individually training the architecture for several hours. Moreover, we further propose a latent representation sampling technique 
to enhance the correlations of pseudo labels to true performance labels. 

Third, since performance metrics and their distributions differ by task, we use several scaling techniques such as standardization and FLOPs-based transform to rescale labels when adapting AIO-P to a downstream task. AIO-P learns a unitless understanding of architecture performance that can then revert into a task-appropriate metric like Average Precision (AP) or mean Intersection over Union (mIoU). 

Through extensive experiments, we demonstrate that AIO-P pretrained on one or two tasks in addition to classification is able to transfer to predict neural network performance for a diverse range of CV tasks including 2D Human Pose Estimation (HPE), Object Detection (OD), Instance Segmentation (IS), Semantic Segmentation (SS), and Panoptic Segmentation (PS). AIO-P consistently achieves a Mean Absolute Error (MAE) below 1\% on task-specific metrics and a Spearman's Rank Correlation Coefficient (SRCC) above 0.5, outperforming both ZCP and GNN baselines under zero-shot transfer and minimum fine-tuning settings. In addition, AIO-P is able to correctly rank architectures in foreign model zoos whose body networks are different from those observed in training, including DeepLab~\cite{deeplabv3plus2018} and TF-Slim~\cite{TFSlim}. Finally, by pairing AIO-P with a search algorithm, we can optimize a proprietary facial recognition model to preserve performance while reducing FLOPs by over 13.5\%. We open-source\footnote{https://github.com/Ascend-Research/AIO-P} our data, code, and predictor design to advance research in this field.
\section{Related Work}
\label{sec:related}

Benchmark datasets and neural predictors provide a quick avenue for performance estimation. Arguably, the most significant difference is how performance is queried. For example, NAS-Bench-101 and 201 contain 423k and 15.6k architectures, respectively. They store individual architecture performances on a look-up table. By contrast, NAS-Bench-301 operates on the DARTS search space, which contains $~10^{18}$ architectures which are too many to evaluate individually. Instead, they train a neural predictor~\cite{zela2022surrogate, luo2020semi, lu2021TNASP, wu2021stronger}. In both cases, architecture configurations define the keys to the table or predictor input, and serve to showcase the limitations of these approaches. Specifically, these configurations are for micro, cell-based NAS, where a network is built by stacking identical cell structures. These configurations generally assume details like latent representation sizes and the number of cells in the network to be constant.

On the other hand, OFA networks~\cite{cai2020once} use macro search spaces where networks are built by individually selecting and then stacking pre-defined blocks. It searches over the number of blocks in the network, kernel size and channel expansions~\cite{mills2021profiling} within a block. Regardless, both approaches assume the stem and head of the network to be a nonsearchable fixed structure. This is acceptable when considering one task. They also abstract multiple operations into pre-defined sequences, e.g., MBConv block, that is specific to the search space and network body and thus may not exist in the head. By contrast, we aim to predict performance across various tasks with different heads and data sizes. Thus, we require a more robust and generalizable data format. AIO-P uses Compute Graphs (CG) as input, which incorporates full network topological details,  latent tensors size, and node features.

A few approaches appear in the literature regarding NAS for CV tasks other than IC. For example, \citet{ding2021learning} use NAS on nine tasks, including IC and SS. They perform a search to find architectures that provide high-performance on multiple objectives. Auto-DeepLab~\cite{liu2019auto} reconfigure DARTS to perform variable upsampling/downsampling and search for a good SS architecture. Next, TransNAS-Bench-101~\cite{duan2021transnas} train architectures from separate micro and macro search spaces on various tasks for several hours and demonstrate the performance of multiple search algorithms. In contrast, we train architectures using the hyperparameters of Detectron2~\cite{wu2019detectron2} and \citet{Zhou_2017_ICCV}. We develop a novel shared head approach to build a generalizable performance predictor that learns on data from multiple CV tasks and predicts the performance of other downstream tasks.
\section{Methodology}
\label{sec:methodology}

\begin{figure}
	\centering
	\includegraphics[width=3.25in]{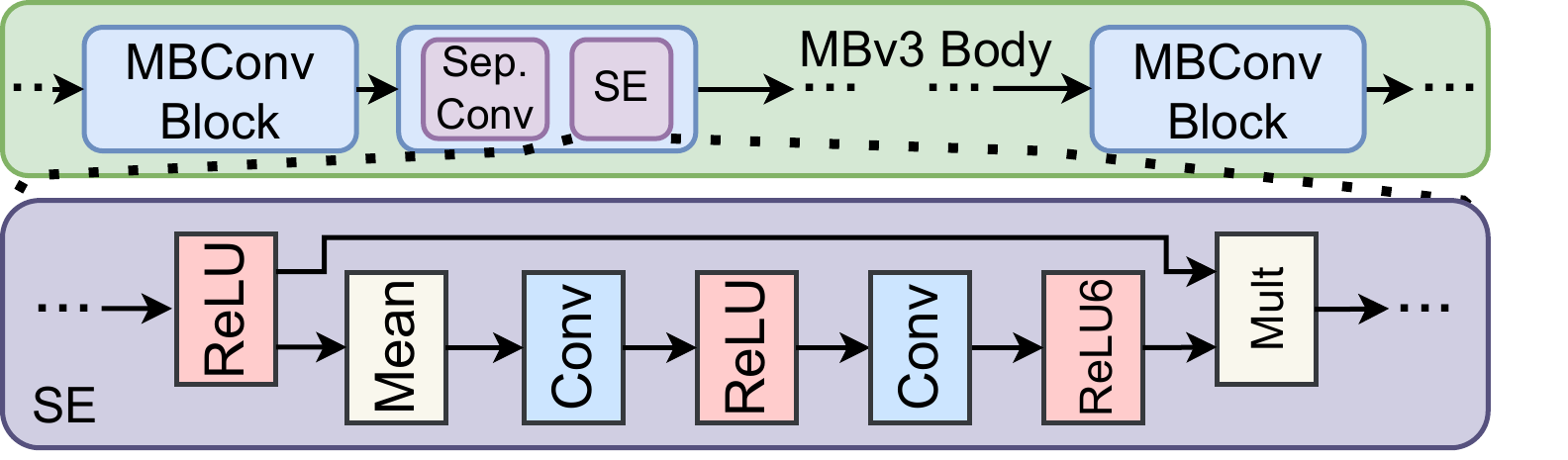}
	\caption{Example CG subgraph of a Squeeze-and-Excite (SE) module~\cite{hu2018squeeze} in MBv3. We store properties like kernel size and channels as node features.} 
	\label{fig:cg_se}
\end{figure}

We cast performance prediction as a supervised learning regression problem where each data sample belongs to a given task domain $t$. Each task $t$ contains instances $(A_i^t, y_i^t)$ where $A_i$ is an architecture, and $y_i$ is its performance label denoted by a metric value on $t$. For example, $y_i^t$ is the \textit{accuracy} value for an Image Classification (IC) architecture, or \textit{average precision} for Object Detection (OD). The goal is a neural predictor that can generalize across many tasks and provide accurate performance estimations. In this section, we describe how AIO-P represents architectures as Computational Graphs and uses $K$-Adapters to learn task transferable knowledge. Furthermore, we describe how shared task heads and pseudo-labeling let us form a large dataset of training instances. Finally, we discuss label scaling techniques for making accurate predictions across different performance distributions.

\subsection{Network Representation}
A CV architecture usually consists of a `body' that performs feature extraction on an input and a `head' that uses extracted features to make predictions. The `body' structure comes from a given search space while the head comes from a specified task $t$. 

The search spaces we consider are from OFA. These include ProxylessNAS (PN)~\cite{cai2018proxylessnas}, MobileNetV3 (MBv3)~\cite{howard2019searching} and ResNet-50 (R50)~\cite{he2016deep}. Architectures bodies from these spaces are pre-trained on ImageNet classification. Head designs vary with task complexity. For example, a typical IC head uses global average pooling and MLP layers to predict class labels for an entire image. By contrast, we consider tasks like HPE, OD, and segmentation, which upsample feature maps with different resolutions to make predictions. Specifically, HPE~\cite{zheng2020deep} generates large heat maps to estimate joint locations, while Semantic Segmentation (SS)~\cite{liu2019auto} upsamples to predict class labels for every pixel in an image. OD~\cite{ren2015faster} uses a Region Proposal Network (RPN) and Region of Interest Poolers (RoI) to estimate bounding box coordinates and classes, while Instance Segmentation (IS) replaces the bounding boxes with pixel-level masks for each instance of a class. Finally, Panoptic Segmentation (PS)~\cite{kirillov2019panoptic} combines IS and SS to generate pixel masks for an image while differentiating individual instances of each class, e.g., masks for every person in a crowd will have different colors. We illustrate this breakdown of how search space bodies pair and interact with task heads in Figure~\ref{fig:head_and_body} and provide finer details for each task head in the supplementary materials. 

While prior neural predictors like \citet{white2019bananas} as well as \citet{wen2020neural} adopt a customized encoding limited to predefined search spaces, e.g., NAS-Benchmarks on IC, we use a general encoding that can represent any neural network. Specifically, we derive Computational Graphs (CG) using the underlying graph structure that libraries like TensorFlow~\cite{tensorflow2015-whitepaper} generate in the forward pass to execute backpropagation. 

CGs are fine-grained graph structures where nodes denote individual atomic operations such as convolutions, linear layers, pooling, padding, addition, concatenation, etc. Node features describe properties like kernel sizes, strides, and channels. Featureless, directed edges denote the flow of latent data across the network, allowing for a task-transferable representation that incorporates the body and task head. Figure~\ref{fig:cg_se} illustrates how CGs represent one module of a MobileNetV3 (MBv3) body. 

Finally, we note that our CG format can be extended to other network types, like those which perform Natural Language Processing tasks such as recurrent or attention-based models or even simple MLP models. However, such experiments are beyond the scope of this paper.

\begin{figure}
	\centering
	\includegraphics[width=3.25in]{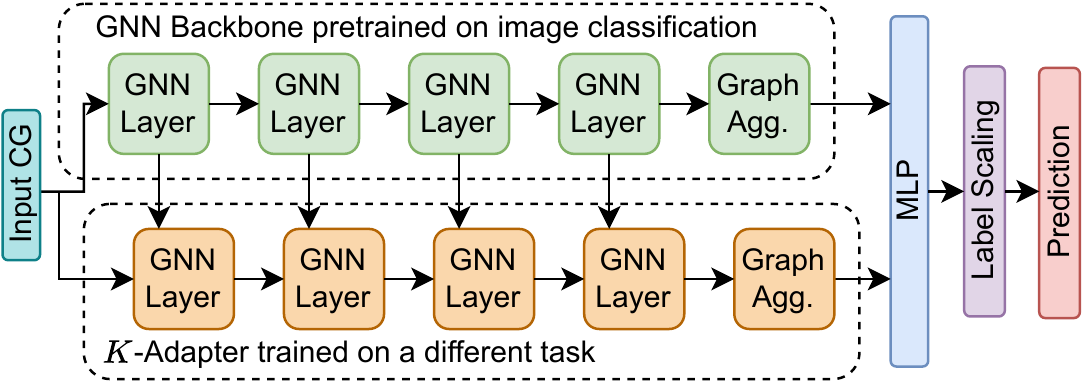}
	\caption{CG K-Adapter Diagram. We start with a graph encoder pre-trained on NAS-Bench-101 for IC and further extend the design with an adapter. The original encoder is frozen while we train the K-Adapter on a new search space and task, e.g., R50 on OD.}
	\label{fig:k_adapt}
\end{figure}

\subsection{$K$-Adapters for Knowledge Infusion}
\label{sec:k_adapt}

AIO-P, as shown in Figure~\ref{fig:k_adapt}, starts with a GNN~\cite{morris2019weisfeiler} regressor backbone. The stem consists of an encoding layer that transforms discrete node features, e.g., operation categories, into a continuous format. The GNN encoder takes the computational graph as input, learns node embeddings using the adjacency matrix, and generates an overall graph embedding by aggregating node embeddings. The graph embedding is fed into an MLP, which outputs a performance prediction. In all our experiments, we pre-train the GNN regressor to predict classification accuracy on 50k NAS-Bench-101 CGs.

To predict performance on another downstream CV task, we extend this base GNN regressor with $K$-Adapters to infuse external knowledge from other tasks and datasets beyond image classification. To do this, we discard the original regressor MLP and freeze all remaining weights. We append the $K$-Adapter by pairing each existing GNN layer in the encoder with a new GNN ``Adapter" layer. Each adapter layer accepts concatenated input from the previous adapter layer and the adjacent GNN layer. Formally, given the intermediate node embeddings for a graph $x$, we define the forward function to $K$-Adapter layer $GNN^{k,i}$ as
\begin{equation}
    \centering
    \label{eq:k_adapt_input}
    h_x^{k,i} = GNN^{k,i}(Concat[h_x^{k,i-1}, h_x^{b,i}]),
\end{equation}
where $h_x^{k,i-1}$ and $h_x^{b,i}$ are the intermediate node embeddings produced by the previous $K$-Adapter layer and adjacent GNN layer in the backbone, respectively. Like the backbone, the $K$-Adapter produces an overall graph embedding. We concatenate both graph embeddings and feed them into a new MLP predictor.

Note that we can augment the original backbone with multiple $K$-Adapters, where each $K$-Adapter infuses the knowledge from a different task $t$. Hence, the overall predictor can generalize to different downstream tasks with different task head CG structures and labels.

\begin{table}[t]
    \centering
    \scalebox{0.79}{
    \begin{tabular}{l|c|c|c} \toprule
    &  \textbf{PN} & \textbf{MBv3} & \textbf{R50} \\ \midrule
    \#Architectures & 215 & 217 & 215 \\ 
    Ground-Truth & 65.16 $\pm$ 0.80\% & 65.22 $\pm$ 0.77\% & 65.64 $\pm$ 0.88\% \\ \midrule
    Body Swap PCK & 52.67 $\pm$ 1.47\% & 43.62 $\pm$ 0.58\% & 49.61 $\pm$ 0.56\% \\
    SRCC & 0.574 & 0.443 & 0.246 \\ \midrule
    Sampling PCK & 59.57 $\pm$ 1.06\% & 61.58 $\pm$ 0.99\% & 59.46 $\pm$ 6.08\% \\
    SRCC & \textbf{0.659} & \textbf{0.576} & \textbf{0.375} \\ 
    \bottomrule
    \end{tabular}
    }
    \caption{Shared head performance distributions and SRCC on HPE PCK [\%]. We compare performance obtained using body swapping and latent sampling to the ground-truth PCK from individually training architectures.}
    \label{tab:sample_srcc}
\end{table}

\subsection{Training $K$-Adapters based on Latent Sampling}
\label{sec:shared_head} 

To train $K$-Adapters, we need datasets composed of architecture samples and their performance labels on tasks other than IC. We can obtain task-specific labels by sampling a body from a search space, e.g., OFA, coupling it with a task head, and training it. However, this is costly. Rather than having AIO-P learn on ground-truth labels obtained from individually trained architectures, we propose a pseudo-labeling method to efficiently obtain task-specific labels to train adapters, via a specially-trained task head that is \textit{shared} amongst all network bodies in a search space.

A typical approach for training a shared task head is `body swapping', which involves iteratively sampling pre-trained architecture bodies from a search space, e.g., OFA, attaching them to the shared head and training the head on a few batches of images. However, note that during body swapping, for each mini-batch of images, the head only samples a single body. To encourage randomness, rather than sampling body networks, we propose an approach where we directly sample the latent representations of the mini-batch of images from a distribution as if they are generated by sampled bodies. Yet, this will further mimic random body sampling per image instead of per mini-batch.  

Specifically, let $x$ be an image, $\mathcal{S}$ be an architecture search space and  $B \in \mathcal{S}$ be a body network. As OFA search spaces each contain roughly $10^{18}$ bodies, we take a subset of architectures $\mathcal{S}'$ and compute the mean $\vec{\mu}(x)$ and standard deviation $\vec{\sigma}(x)$ of their latent representations of image $x$, i.e., 
\begin{equation}
    \centering
    \label{eq:math_mu}
    \vec{\mu}(x) = \mathbb{E}_{B \in \mathcal{S}'}[f_{B}(x)],
\end{equation}
\begin{equation}
    \centering
    \label{eq:math_sig}
    \vec{\sigma}^2(x) = \text{Var}_{B \in \mathcal{S}'}[f_{B}(x)],
\end{equation}
where $f_B$ denotes the function given by network body $B$. We then sample a latent representation $\vec{z}(x) = \vec{\mu}(x) + \vec{\zeta} \circ \vec{\sigma}(x)$ where $\vec{\zeta}$ is a standard normal $\mathcal{N}(0, 1)$ random vector and $\circ$ is element-wise multiplication. We denote this approach as `latent sampling', where we train the shared task head by $\{(\vec{z}(x), y_x)\}$, using the sampled latent vector $\vec{z}(x)$ as input to the head and $y_x$ as the ground-truth task label for $x$. We limit the size of $\mathcal{S}'$ and use a round robin strategy with binning to constantly swap new bodies into $\mathcal{S}'$. We provide procedural details on our round robin strategy, shared head hyperparameters and resource cost breakdown in the supplementary materials.

To assess the reliability of our pseudo-labelling approach, we form a ground-truth set by individually training several hundred architectures (with bodies sampled from three spaces) on the Leeds Sports Pose-Extended (LSP) dataset for Human Pose Estimation (HPE) and measure performance in terms of Percentage of Correct Keypoints (PCK). We obtain pseudo-labels by fine-tuning the same bodies when connected to shared heads trained using \textit{body swapping} and \textit{latent sampling}, respectively. We compare the PCK distributions and SRCC between the pseudo-labels and ground-truth labels. We do not use the pseudo-labeled architectures from this experiment to train AIO-P. They are only for comparison. Table~\ref{tab:sample_srcc} shows the results. We note that latent sampling achieves much better performance on average, relative to the ground truth, while body swapping lags by over 10\%. Also, the SRCC we observe using the latent sampling approach is positive for all search spaces and more favorable than body swapping.

\begin{table}[t]
    \centering
    \scalebox{0.79}{
    \begin{tabular}{l|c|c|c} \toprule
    \textbf{Task} & PN & MBv3 & R50  \\ \midrule
    HPE-LSP    & 215/2580 &217/2862  &215/2986 \\
    HPE-MPII   & 246/- &236/-  &236/- \\
    OD/IS/SS/PS & 118/1633 &118/1349  &115/1399 \\ \bottomrule
    \end{tabular}
    }
    \caption{AIO-P $K$-Adapter dataset size, across each search space and task. `/' denotes architectures that we individually train (left) and those we label using a shared head (right).}
    \label{tab:dataset_size}
\end{table}

\begin{table*}[t]
    \centering
    \scalebox{0.75}{
    \begin{tabular}{l|ccc|ccc|ccc} \toprule
    &  & \textbf{ProxylessNAS} &  &  & \textbf{MobileNetV3} & & & \textbf{ResNet-50} &  \\ \midrule
    \textbf{Task} & GNN & +Eqs.~\ref{eq:standardization} \& \ref{eq:accTrans} & AIO-P &  GNN & +Eqs.~\ref{eq:standardization} \& \ref{eq:accTrans} & AIO-P &  GNN & +Eqs.~\ref{eq:standardization} \& \ref{eq:accTrans} & AIO-P \\ \midrule
    LSP  & 27.27 $\pm$ 0.39\% & 0.72 $\pm$ 0.14\%&\textbf{0.70} $\pm$ 0.23\% & 27.07 $\pm$ 0.53\% &0.74 $\pm$ 0.11\%& \textbf{0.52} $\pm$ 0.01\% & 21.47 $\pm$ 2.56\% & 1.07 $\pm$ 0.07\% &\textbf{0.94} $\pm$ 0.05\% \\
    MPII & 8.10 $\pm$ 0.38\%  & \textbf{0.34} $\pm$ 0.07\%& 0.42 $\pm$ 0.13\% &  8.91 $\pm$ 0.52\%&0.36 $\pm$ 0.05\%& \textbf{0.27} $\pm$ 0.01\%  & 2.48 $\pm$ 1.93\% & 1.11 $\pm$ 0.05\% &\textbf{1.03} $\pm$ 0.07\% \\
    OD   & 59.53 $\pm$ 0.41\% & 1.15 $\pm$ 0.46\%&\textbf{0.63} $\pm$ 0.09\% & 59.56 $\pm$ 0.37\% &1.24 $\pm$ 0.14\%& \textbf{0.62} $\pm$ 0.04\% & 54.07 $\pm$ 1.31\% & 0.88 $\pm$ 0.18\% &\textbf{0.60} $\pm$ 0.02\% \\
    IS   & 62.00 $\pm$ 0.34\% & 0.93 $\pm$ 0.18\%&\textbf{0.52} $\pm$ 0.14\% & 61.76 $\pm$ 0.33\% &0.73 $\pm$ 0.07\%& \textbf{0.58} $\pm$ 0.07\% & 56.74 $\pm$ 1.66\% & 0.64 $\pm$ 0.09\% &\textbf{0.48} $\pm$ 0.01\% \\
    SS   & 53.07 $\pm$ 0.37\% & 0.71 $\pm$ 0.22\%&\textbf{0.50} $\pm$ 0.06\% & 53.40 $\pm$ 0.39\% &0.90 $\pm$ 0.09\%& \textbf{0.56} $\pm$ 0.04\% & 47.90 $\pm$ 2.68\% & 0.52 $\pm$ 0.02\% &\textbf{0.44} $\pm$ 0.02\% \\
    PS   & 56.19 $\pm$ 0.35\% & 0.76 $\pm$ 0.07\%&\textbf{0.50} $\pm$ 0.10\% & 56.18 $\pm$ 0.33\% &0.74 $\pm$ 0.05\%& \textbf{0.57} $\pm$ 0.04\% & 51.63 $\pm$ 1.69\% & 0.52 $\pm$ 0.04\% &\textbf{0.45} $\pm$ 0.01\% \\ \bottomrule
    \end{tabular}
    }
    \caption{MAE [\%] of AIO-P on three search spaces and six tasks in the zero-shot transfer setting (no fine-tuning), compared to 
    GNN without and with rescaling by Eqs.~\ref{eq:standardization} \& \ref{eq:accTrans}. AIO-P adopts 2 $K$-Adapters, trained on LSP and OD. AIO-P uses Equation~\ref{eq:accTrans} and standardizes regression targets. Results averaged across 5 seeds. }
    \label{tab:main_mae}
\end{table*}

\begin{table*}[t]
    \centering
    \scalebox{0.75}{
    \begin{tabular}{l|ccc|ccc|ccc} \toprule
    &  & \textbf{ProxylessNAS} &  &  & \textbf{MobileNetV3} & & & \textbf{ResNet-50} &  \\ \midrule
    \textbf{Task} & GNN & +Eqs.~\ref{eq:standardization} \& \ref{eq:accTrans} & AIO-P &  GNN & +Eqs.~\ref{eq:standardization} \& \ref{eq:accTrans} & AIO-P &  GNN & +Eqs.~\ref{eq:standardization} \& \ref{eq:accTrans} & AIO-P \\ \midrule
    LSP  & 0.593 $\pm$ 0.02 & 0.561 $\pm$ 0.05&\textbf{0.698} $\pm$ 0.01 & 0.259 $\pm$ 0.09 & 0.418 $\pm$ 0.06&\textbf{0.556} $\pm$ 0.01 & -0.302 $\pm$ 0.02 & 0.176 $\pm$ 0.03&\textbf{0.261} $\pm$ 0.02 \\
    MPII & 0.711 $\pm$ 0.01 & \textbf{0.767} $\pm$ 0.02&0.753 $\pm$ 0.01 & 0.300 $\pm$ 0.14 & \textbf{0.764} $\pm$ 0.01&0.701 $\pm$ 0.02 & -0.315 $\pm$ 0.02 & 0.446 $\pm$ 0.03&\textbf{0.532} $\pm$ 0.02 \\
    OD   & 0.558 $\pm$ 0.06 & 0.471 $\pm$ 0.11&\textbf{0.781} $\pm$ 0.03 & \textbf{0.645} $\pm$ 0.05 & 0.087 $\pm$ 0.14& 0.515 $\pm$ 0.05 & -0.489 $\pm$ 0.11 & 0.645 $\pm$ 0.03&\textbf{0.817} $\pm$ 0.01 \\ 
    IS   & 0.599 $\pm$ 0.07 & 0.211 $\pm$ 0.10&\textbf{0.831} $\pm$ 0.02 & 0.592 $\pm$ 0.07 & 0.034 $\pm$ 0.03&\textbf{0.602} $\pm$ 0.05 & -0.493 $\pm$ 0.08 & 0.495 $\pm$ 0.04&\textbf{0.817} $\pm$ 0.01 \\ 
    SS   & 0.487 $\pm$ 0.03 & 0.262 $\pm$ 0.18&\textbf{0.735} $\pm$ 0.02 & 0.517 $\pm$ 0.08 & -0.367 $\pm$ 0.11& \textbf{0.689} $\pm$ 0.02 & -0.406 $\pm$ 0.03 & 0.589 $\pm$ 0.03&\textbf{0.660} $\pm$ 0.02 \\ 
    PS   & 0.562 $\pm$ 0.00 & 0.119 $\pm$ 0.12&\textbf{0.732} $\pm$ 0.03 & \textbf{0.570} $\pm$ 0.07 & -0.009 $\pm$ 0.06& 0.518 $\pm$ 0.04 & -0.455 $\pm$ 0.07 & 0.599 $\pm$ 0.03& \textbf{0.788} $\pm$ 0.02 \\ \bottomrule
    \end{tabular}
    }
    \caption{SRCC of AIO-P on three search spaces and six tasks. Same configurations as Table~\ref{tab:main_mae}.}
    \label{tab:main_srcc}
\end{table*}

\subsection{Label Scaling}
\label{sec:label_scaling}

In addition to having different heads, task performance distributions differ. PN architectures yield $\sim$75\% accuracy on ImageNet, and around 40\% SS mIoU on COCO. Note that performance distributions differ amongst tasks, or even the same task between ground truth and pseudo-labels.

Therefore, we experiment with methods that scale the labels AIO-P learns from to add further task transferability. Specifically, we employ standardization, 

\begin{equation}
    \centering
    \label{eq:standardization}
    \mathcal{Z}(y) = \dfrac{y - \mu}{\sigma},
\end{equation}
where $\mu$ and $\sigma$ are the mean and standard deviation of the label $y$ distribution, respectively. This approach fits a set of data into a normal distribution $\mathcal{N}(0, 1)$. Further, we incorporate scaling by FLOPs, or Floating Point Operations required to perform the forward pass of a network, as a divisor prior to standardization, 

\begin{equation}
    \centering
    \label{eq:accTrans}
    y_{F} = y\cdot(\texttt{Log}_{10}(F+1)+1)^{-1},
\end{equation}
where $F$ are the FLOPs of the network with performance $y$, measured in GigaFLOPs (1e9), and the addition of 1 in the denominator assures it will be positive real number. In all cases, if we apply Equation~\ref{eq:accTrans} to labels, we then standardize them using Equation~\ref{eq:standardization}.  

\section{Results}
\label{sec:results}

\begin{table*}[t]
    \centering
    \scalebox{0.8}{
    \begin{tabular}{l|c|c|c|c|c|c|c|c} \toprule
    \textbf{Space} & \textbf{Synflow}  & \textbf{Jacov} & \textbf{Fisher} & \textbf{Gradient Norm} & \textbf{Snip} & \textbf{FLOPs} & \textbf{AIO-P} & \textbf{AIO-P FT}\\ \midrule
    PN-LSP     & -0.004 $\pm$ 0.03 & -0.057 $\pm$ 0.04& 0.449 $\pm$ 0.03& 0.581 $\pm$ 0.03 &0.624 $\pm$ 0.16& 0.584 $\pm$ 0.01& \textbf{0.698} $\pm$ 0.01& \textit{0.668} $\pm$ 0.03\\
    MBv3-LSP   & \textbf{0.609} $\pm$ 0.01& 0.029 $\pm$ 0.10&0.129 $\pm$ 0.06& 0.426 $\pm$ 0.03 & 0.466 $\pm$ 0.01& 0.562 $\pm$ 0.01& 0.556 $\pm$ 0.01& \textit{0.567} $\pm$ 0.01\\
    R50-LSP    & \textbf{0.639} $\pm$ 0.01& -0.071 $\pm$ 0.03&0.515 $\pm$ 0.01& 0.581 $\pm$ 0.02 & \textit{0.646} $\pm$ 0.02& 0.263 $\pm$ 0.02& 0.261 $\pm$ 0.02& 0.264 $\pm$ 0.02\\ \midrule \midrule
    PN-MPII    & 0.046 $\pm$ 0.07& -0.008 $\pm$ 0.08&0.538 $\pm$ 0.03& 0.733 $\pm$ 0.01 & \textit{0.755} $\pm$ 0.01 & 0.735 $\pm$ 0.00& 0.753 $\pm$  0.01& \textbf{0.773} $\pm$ 0.02\\
    MBv3-MPII  & \textit{0.736} $\pm$ 0.01& -0.014 $\pm$ 0.09&0.203 $\pm$ 0.08& 0.679 $\pm$ 0.03 & 0.691 $\pm$ 0.04& \textit{0.736} $\pm$ 0.01& 0.701 $\pm$ 0.02& \textbf{0.744} $\pm$ 0.05\\
    R50-MPII   & \textbf{0.865} $\pm$ 0.01& 0.111 $\pm$ 0.08&0.709 $\pm$ 0.02& 0.732 $\pm$ 0.02 & \textit{0.849} $\pm$ 0.02& 0.532 $\pm$ 0.02& 0.532 $\pm$ 0.02& 0.532 $\pm$ 0.02\\ \midrule \midrule
    PN-SS      & 0.022 $\pm$ 0.07& -0.023 $\pm$ 0.13& 0.050 $\pm$ 0.07& 0.141 $\pm$ 0.06 & -0.082 $\pm$ 0.07& 0.608 $\pm$ 0.01& \textit{0.735} $\pm$ 0.02& \textbf{0.849} $\pm$ 0.03\\
    MBv3-SS    & -0.309 $\pm$ 0.07& 0.042 $\pm$ 0.08&0.022 $\pm$ 0.06& 0.040 $\pm$ 0.06 &0.188 $\pm$ 0.04& 0.445 $\pm$ 0.02& \textit{0.689} $\pm$ 0.02& \textbf{0.822} $\pm$ 0.03\\
    R50-SS     & -0.255 $\pm$ 0.09& 0.141 $\pm$ 0.10&0.126 $\pm$ 0.06& 0.354 $\pm$ 0.08 & 0.036 $\pm$ 0.07& \textit{0.661} $\pm$ 0.02& 0.660 $\pm$ 0.02& \textbf{0.677} $\pm$ 0.03\\ \bottomrule
    \end{tabular}
    }
    \caption{SRCC between AIO-P, ZCP methods and a FLOPs-based predictor on three tasks. We also include `AIO-P FT', i.e., fine-tuning on 20 held-out standardization samples. We bold the best result and italicize the second best.}
    \label{tab:zcp}
    \vspace{-2mm}
\end{table*}

In this Section, we describe our suite of tasks, experimental setup and results. We consider the computer vision tasks of 2D Human Pose Estimation (HPE), Object Detection (OD), Instance Segmentation (IS), Semantic Segmentation (SS) and Panoptic Segmentation (PS).

HPE predicts joint locations from an image. We measure 2D HPE performance using Percentage of Correct Keypoints (PCK) and consider two HPE datasets: MPII~\cite{MPII} and Leeds Sports Pose-Extended (LSP)~\cite{lsp_extended}, which contain 22k and 11k images, respectively. We individually train networks for both datasets and train a shared head for LSP.

OD and IS measure performance in mean Average Precision (mAP), while SS uses mean Intersection over Union (mIoU). PS, a combination of IS and SS, uses Panoptic Quality (PQ), a balance of mAP and mIoU. These performance metrics are reported as percentages [\%]. We consider the 2017 version of MS Common Objects in Context (COCO)~\cite{coco} as our dataset for these tasks, as it contains 118k and 5k training and validation images, respectively. We use Detectron2~\cite{wu2019detectron2} to pair a given body with OD, IS SS, and PS to train on all four tasks simultaneously. 

Table~\ref{tab:dataset_size} enumerates of the number of architectures we have for each search space and task. We use pseudo-labeled architecture CGs to train AIO-P and reserve individually trained ones as held-out test samples. Additionally, we include a more extensive breakdown with performance and FLOPs distributions in the supplementary materials. 

\subsection{Training Procedure}
\label{sec:training}

Starting with the GNN backbone, we append two $K$-Adapter modules. We train these using the pseudo-labeled CGs for two task domains, OD and LSP-HPE. We separately apply Equation~\ref{eq:accTrans} and then standardize the labels of each $K$-Adapter task before training and freeze the weights of the GNN backbone. 

To evaluate individually trained test set CGs for a task, we consider the zero-shot transfer context where we provide no data on a target task prior to inference. Similar to \citet{lu2021one}, we sample 20 random architectures from the test set to compute standardization parameters $\mu$ and $\sigma$ and exclude these architectures from evaluation. We also consider a fine-tuning context where we use the same 20 random architectures to train AIO-P prior to inference. As we evaluate multiple methods using different random seeds, we ensure the set of 20 sampled architectures is the same for every seed value. Also, note that the set of body architectures we individually train on any task are disjoint from the ones we pseudo-label. We provide granular predictor details and training hyperparameters in the supplementary materials.

\subsection{Zero-Shot Transfer Performance}
\label{sec:zero_shot}

We consider two predictor evaluation metrics, Mean Absolute Error (MAE) and Spearman's Rank Correlation Coefficient (SRCC). The former gauges a predictor's ability to make accurate estimations on individual labels, while the latter determines a predictor's capacity to correctly rank a population by performance. We report MAE as a percentage for each task metric, e.g., PCK for HPE and mIoU for SS, where lower is better. SRCC falls in $[-1, 1]$ and indicates agreement with ground-truth ordering, so higher is better. 

Tables~\ref{tab:main_mae} and \ref{tab:main_srcc} list our results for AIO-P in terms of MAE and SRCC, respectively. AIO-P achieves the best MAE performance in the majority of scenarios, including every task on MBv3 and R50. The sole exception is a GNN using standardization and FLOPs scaling for MPII by less than 0.1\% MAE, while the unmodified GNN fails to generalize to different task performance distributions. 

For SRCC, AIO-P consistently achieves the best correlation on all R50 tasks and in at least half for PN and MBv3. While both variants of the GNN obtain the high SRCC on at least one search space and task pair, they are overall very inconsistent as the GNN with re-scaled labels achieves negative SRCC on MBv3 and the unmodified GNN fails on R50.

Next, Table~\ref{tab:zcp} compares the ranking performance of AIO-P with Zero-Cost Proxies (ZCP) and FLOPs. Without fine-tuning, AIO-P achieves high correlation performance for all search spaces on MPII and SS, as well as PN and MBv3 for LSP, demonstrating high generalizability. The most competitive ZCP method is Synflow~\cite{tanaka2020pruning}, but only for HPE tasks with MBv3 and R50 as it fails on PN and in the SS context. Grad Norm achieves positive SRCC across all search spaces and tasks but never the best performance. FLOPs are also highly correlated with performance in all settings, except LSP for R50. Unlike ZCP, the advantage of AIO-P in this context is its ability undergo fine-tuning to boost performance, as shown for the MPII and SS tasks. Next, we compare Eqs.~\ref{eq:standardization} and \ref{eq:accTrans} with another fine-tuning-based method for intertask prediction.

\begin{table}[t]
    \centering
    \scalebox{0.8}{
    \begin{tabular}{l|c|c|c|c} \toprule
    \textbf{Task} & \textbf{GNN} & \textbf{AIO-P+AdaProxy} & \textbf{AIO-P w/o Eq.~\ref{eq:accTrans}} & \textbf{AIO-P} \\ \midrule 
    MPII       & 0.33\% & 0.35\% & \textit{0.29}\% & \textbf{0.25}\%  \\
    Inst. Seg. & 0.66\% & \textit{0.47}\% & \textbf{0.39}\% & 0.58\% \\
    Pan. Seg.  & 0.64\% & \textit{0.51}\% & \textbf{0.46}\% & 0.61\% \\ \bottomrule
    \end{tabular}
    }
    \caption{MAE of AIO-P on MPII, IS and PS for the MBv3 search space (with fine-tuning). A single $K$-Adapter was trained on SS for AIO-P variants. We bold the best result and italicize the second best.}
    \label{tab:rescale_mae}
\end{table}

\begin{table}[t]
    \centering
    \scalebox{0.8}{
    \begin{tabular}{l|c|c|c|c} \toprule
    \textbf{Task} & \textbf{GNN} & \textbf{AIO-P+AdaProxy} & \textbf{AIO-P w/o Eq.~\ref{eq:accTrans}} & \textbf{AIO-P} \\ \midrule 
    MPII       & \textit{0.680} & 0.574 & 0.650 & \textbf{0.750}  \\
    Inst. Seg. & 0.547 & \textit{0.677} & \textbf{0.747} & 0.585  \\
    Pan. Seg.  & 0.568 & \textit{0.627} & \textbf{0.685} & 0.512  \\ \bottomrule
    \end{tabular}
    }
    \caption{SRCC of AIO-P on MPII, IS and PS for the MBv3 search space (with fine-tuning). Same experimental setup as Table~\ref{tab:rescale_mae}.}
    \label{tab:rescale_srcc}
\end{table}

\subsection{Fine-Tuning Results}
\label{sec:adaproxy}

We now evaluate the transferability of AIO-P on downstream tasks if predictor fine-tuning is allowed based on a small number, i.e., 20, downstream architectures. As a comparison, we also evaluate AIO-P using a weight scaling technique proposed in AdaProxy~\cite{lu2021one}, which scales the final MLP layer of a predictor by minimizing the following loss on fine-tuning samples:
\[\min_{\alpha, \vec{b}} |[(\alpha\vec{I}^T + \vec{b}^T) \circ \vec{w}^T]\vec{x} - y|^{2} + \lambda|\vec{b}|,\]
where $\alpha$ is a scalar, $\vec{I}$ is an identity vector, $\vec{w}$ are the weights in the final layer, $\vec{b}$ is a sparsity vector and $\lambda$ is a regularizer weight. During this minimization, all of the original predictor weights are frozen. For this experiment, we consider the MBv3 search space, train a $K$-Adapter on SS and evaluate on MPII, IS and PS. 

Tables~\ref{tab:rescale_mae} and \ref{tab:rescale_srcc} list our results in terms of MAE and SRCC, respectively. We are able to obtain more accurate predictions in terms of MAE and SRCC using some form of standardization rather than AdaProxy. On MPII specifically, Eq.~\ref{eq:accTrans} overcomes the limitations of low inter-task correlation to produce the best MAE and SRCC, while just using standardization is enough to obtain the most accurate IS and PS predictions. Another advantage of our standardization and FLOPs-based transform approach is the absence of tunable hyperparameters, e.g., $\lambda$. 

\begin{table}[t]
    \centering
    \scalebox{0.8}{
    \begin{tabular}{l|c|c|c|c|c} \toprule
    \textbf{Task} & \textbf{OD} & \textbf{SS} & \textbf{LSP} & \textbf{OD+SS} & \textbf{OD+LSP}  \\ \midrule
    MPII    & 0.60\% & 0.39\% & \textbf{0.28}\% & \textit{0.35}\% & 0.42\% \\
    IS      & 0.70\% & \textit{0.56}\% & 1.03\% & 0.61\% & \textbf{0.52}\% \\
    PS      & \textit{0.75}\% & 0.82\% & 1.03\% & 0.81\% & \textbf{0.50}\% \\ \midrule
    MPII FT & \textbf{0.25}\% & \textit{0.26}\% & 0.27\% & 0.27\% & \textit{0.26}\% \\
    IS FT   & 0.49\% & 0.50\% & \textbf{0.27}\% & 0.50\% & \textit{0.33}\% \\
    PS FT   & 0.50\% & 0.54\% & \textbf{0.31}\% & 0.52\% & \textit{0.33}\% \\ \bottomrule
    \end{tabular}
    }
    \caption{MAE of different $K$-Adapter tasks on the PN search space.     `FT' indicates fine-tuning on 20 held-out target architectures. We bold the best result and italicize the second best.}
    \label{tab:ablation_k_mae}
\end{table}

\begin{table}[t]
    \centering
    \scalebox{0.8}{
    \begin{tabular}{l|c|c|c|c|c} \toprule
    \textbf{Task} & \textbf{OD} & \textbf{SS} & \textbf{LSP} & \textbf{OD+SS} & \textbf{OD+LSP}  \\ \midrule
    MPII    & \textbf{0.762} & 0.660 & 0.717 & 0.727 & \textit{0.753} \\
    IS      & 0.749 & 0.728 & \textbf{0.929} & 0.750 & \textit{0.831} \\
    PS      & 0.650 & 0.669 & \textbf{0.859} & 0.672 & \textit{0.732} \\ \midrule
    MPII FT & \textbf{0.793} & \textit{0.790} & 0.764 & 0.779 & 0.773 \\
    IS FT   & 0.749 & 0.757 & \textbf{0.915} & 0.730 & \textit{0.894} \\
    PS FT   & 0.670 & 0.671 & \textbf{0.880} & 0.647 & \textit{0.858} \\ \bottomrule
    \end{tabular}
    }
    \caption{SRCC of different $K$-Adapter tasks on the PN search space. Same experimental setup as Table~\ref{tab:ablation_k_mae}.}
    \label{tab:ablation_k_srcc}
\end{table}

\subsection{Ablation Study}
\label{sec:ablation}

We ablate the effect of different $K$-Adapter training tasks using PN as an example search space. Tables~\ref{tab:ablation_k_mae} and \ref{tab:ablation_k_srcc} show our MAE and SRCC findings, respectively. We see that using a double $K$-Adapter on OD and LSP helps to generalize MAE and SRCC performance across the downstream tasks. While the best results typically use just OD or LSP, LSP struggles to produce low MAE on IS and PS without fine-tuning. However, we overcome this hurdle when adding another $K$-Adapter for OD. First, we note that the best results use either OD or LSP, although SS still produces good results. Particularly, LSP struggles to produce low MAE on IS and PS without fine-tuning. Introducing another $K$-Adapter for OD overcomes this hurdle. 

Additionally, we compare the GNN backbone to using a single $K$-Adapter, using standardization (Eq.~\ref{eq:standardization}), and then our FLOPs scaling (Eq.~\ref{eq:accTrans}). Tables~\ref{tab:transform_mae} and \ref{tab:transform_srcc} list our results for MAE and SRCC, respectively. We see that AIO-P with or without Eq.~\ref{eq:accTrans} can obtain top MAE performance in most scenarios in the zero-shot and fine-tuning contexts. However, that is not the case for SRCC, where Eq.~\ref{eq:accTrans} allows AIO-P to achieve correlation metrics above 0.8 with fine-tuning. While the normal GNN is not very competitive, adding a single $K$-Adapter without standardization can improve ranking performance significantly. However, the prediction error is still high due to differences in task performance distributions (reported in the Supplementary Materials). Overall, standardization and Eq.~\ref{eq:accTrans} allow AIO-P to strike a balance between prediction error and ranking correlation.

\begin{table}[t!]
    \centering
    \scalebox{0.8}{
    \begin{tabular}{l|c|c|c|c} \toprule
    \textbf{Task} & \textbf{GNN} & \textbf{+$K$-Adapter} & \textbf{+ Eq.~\ref{eq:standardization}} & \textbf{AIO-P} \\ \midrule 
    MPII       & 2.48\% & 55.30\% & \textbf{0.39}\% & \textit{2.38}\%  \\
    IS & 56.74\% & \textit{2.38}\% & \textbf{0.73}\% & \textbf{0.73}\%  \\
    PS  & 51.63\% & 7.39\% & \textit{0.89}\% & \textbf{0.61}\%  \\ \midrule
    MPII-FT & 0.28\% & \textit{0.77}\% & \textbf{0.15}\% & 1.09\%  \\
    IS-FT & 0.61\% & 0.85\% & \textbf{0.36}\% & \textit{0.45}\%  \\
    PS-FT  & 0.71\% & 0.61\% & \textbf{0.40}\% & \textit{0.44}\%  \\ \bottomrule
    \end{tabular}
    }
    \caption{MAE performance comparing the effect of $K$-Adapters as well as Eqs.~\ref{eq:standardization} and \ref{eq:accTrans} in the zero-shot and fine-tuning (FT) contexts. We consider the R50 search space and train a $K$-Adapter on SS. We bold the best result and italicize the second best.}
    \label{tab:transform_mae}
\end{table}

\begin{table}[t]
    \centering
    \scalebox{0.8}{
    \begin{tabular}{l|c|c|c|c} \toprule
    \textbf{Task} & \textbf{GNN} & \textbf{+$K$-Adapter} & \textbf{+ Eq.~\ref{eq:standardization}} & \textbf{AIO-P} \\ \midrule 
    MPII       & -0.315 & \textbf{0.708} & \textit{0.418} & 0.337  \\
    IS & -0.493 & \textbf{0.669} & \textit{0.361} & 0.324  \\
    PS  & -0.455 & \textbf{0.633} & \textit{0.399} & 0.291  \\ \midrule
    MPII-FT       & \textit{0.700} & \textbf{0.738} & 0.512 & 0.532  \\
    IS-FT & 0.611 & 0.687 & \textit{0.727} & \textbf{0.840}  \\
    PS-FT  & 0.601 & 0.667 & \textit{0.762} & \textbf{0.811}  \\ \bottomrule
    \end{tabular}
    }
    \caption{SRCC performance comparing the effect of $K$-Adapters.     Same experimental setup as Table~\ref{tab:transform_mae}.} 
    \label{tab:transform_srcc}
\end{table}

\subsection{Transfer to Foreign Network Types}
\label{sec:model_zoo}


To further test the transferability of AIO-P to new architecture types not seen in training, we perform inference on several foreign `model zoos'. Each model zoo contains 10 or fewer architecture variants, e.g., Inception~\cite{szegedy2017inception}, EfficientNets~\cite{tan2019efficientnet}, MobileNets and ResNets. Specifically, we include model zoos from the DeepLab repository~\cite{deeplabv3plus2018} which focus on 
for semantic segmentation (SS) performance on multiple datasets. We also evaluate AIO-P on several image classification (IC) model zoos from TensorFlow-Slim~\cite{TFSlim}. 

\subsubsection{DeepLab Semantic Segmentation}
We consider architectures on three different SS datasets: ADE20k~\cite{zhou2017ADE}, Pascal VOC~\cite{everingham2015Pascal}, and Cityscapes~\cite{Cordts2016Cityscapes}. These are different SS datasets than the MS-COCO we use for training OFA-based architectures, so the results further demonstrate the generalizability of AIO-P across datasets, even for the same task. Specifically, we consider the following architecture per dataset:

\begin{enumerate}
    \item \textbf{ADE20k} (5): MobileNetV2, Xception65~\cite{chollet2017xception} as well as Auto-DeepLab-\{S, M, L\}~\cite{liu2019auto}.
    \item \textbf{Pascal VOC} (6): MobileNetV2, MobileNetV2 w/ reduced depth, Xception65 and Auto-DeepLab-\{S, M, L\}.
    \item \textbf{Cityscapes} (8): MobileNet\{V2, V3-Small, V3-Large\}, 
    Xception65, Xception71 and Auto-DeepLab-\{S, M, L\}.
\end{enumerate}

\subsubsection{TensorFlow-Slim Image Classification}
We also consider several architectures that perform IC on ImageNet:

\begin{enumerate}
    \item \textbf{ResNets} (6): ResNet-v1-\{50, 101, 152\}~\cite{he2016deep} and ResNet-v2-\{50, 101, 152\}~\cite{he2016identity}.
    \item \textbf{Inception} (5): Inception-\{v1, v2, v3, v4\} and Inception-ResNet-v2~\cite{szegedy2017inception}
    \item \textbf{MobileNets} (6): MobileNet\{V1, V1-0.5, V1-0.25, V2, V2-1.4\} 
    where `-X.Y' is a channel multiplier. 
    \item \textbf{EfficientNets}~\cite{tan2019efficientnet} (8): EfficientNet-\{B0, B1, B2, B3, B4, B5, B6, B7\}.
\end{enumerate}

Using AIO-P pre-trained on ResNet-50 bodies and with a double $K$-Adapter trained on OD and LSP tasks, and then fine-tuned on SS (the task DeepLab performs), we investigate whether AIO-P can adequately rank the architectures in DeepLab and in TensorFlow-Slim, which contain new types of body networks other than ResNet-50. 

As shown in Table~\ref{tab:model_zoo}, we note that AIO-P can achieve positive correlation inference on all SS and IC model zoos. In particular, we obtain perfect SRCC on EfficientNets. Moreover, this performance is superior to AIO-P when Eq.~\ref{eq:accTrans} is not present. This is because while standardization improves MAE performance, it does not affect architecture rankings, whereas FLOPs-based transformation does. These findings demonstrate the efficacy of AIO-P and Eq.~\ref{eq:accTrans} in ranking the performance of foreign networks with different connections/topologies.

\begin{table}[t]
    \centering
    \scalebox{0.8}{
    \begin{tabular}{l|c|c|c} \toprule
    \textbf{Model Zoo} & \textbf{\#Archs} & \textbf{AIO-P w/o Eq.~\ref{eq:accTrans}} & \textbf{AIO-P} \\ \midrule
    DeepLab-ADE20k & 5 & 0.127 $\pm$ 0.255 & \textbf{0.991} $\pm$ 0.016 \\
    DeepLab-Pascal & 6 & 0.392 $\pm$ 0.088 & \textbf{0.939} $\pm$ 0.035 \\
    DeepLab-Cityscapes & 8  & 0.572 $\pm$ 0.031 & \textbf{0.925} $\pm$ 0.024\\ \midrule \midrule
    Slim-ResNets    & 6    & -0.577 $\pm$ 0.183 & \textbf{0.920} $\pm$ 0.106 \\ 
    Slim-Inception  & 5    & -0.700 $\pm$ 0.316 & \textbf{0.980} $\pm$ 0.040 \\
    Slim-MobileNets & 5    & -0.500 $\pm$ 0.000 & \textbf{0.400} $\pm$ 0.535 \\
    Slim-EfficientNets & 8 & \textbf{1.000} $\pm$ 0.000 & \textbf{1.000} $\pm$ 0.000 \\ \bottomrule
    \end{tabular}
    }
    \caption{SRCC of AIO-P on several `model zoos'. Double horizontal line demarcates SS models from IC models.}
    \label{tab:model_zoo}
\end{table}

\subsection{Application to NAS}
\label{sec:nas}

Finally, we apply AIO-P to NAS. Specifically, we use a predictor that achieves high SRCC on the bounding box task OD on R50, to optimize a proprietary neural network designed to perform Facial Recognition (FR) on mobile devices. 
We pair AIO-P with a 
mutation-based search algorithm that aims to preserve 
performance while reducing FLOPs. Although the architecture we optimize does not belong to any of the OFA search spaces we consider, 
like the aforementioned model zoos, AIO-P can estimate performance using the CG framework. Additionally, our mutation algorithm proposes edits to CGs that vary from swapping subgraphs of operation sequences to manually pruning the number of channels in a convolution node. 

Table~\ref{tab:properitary} 
demonstrates that 
we can maintain performance in most settings while reducing FLOPs 
by over 13.5\%. In the `dark' setting, where features can be hard to see, our model improves precision and recall by 0.2\% and 0.6\%, respectively. At most, we only lose 0.2\% precision and 0.8\% recall across other settings. Therefore, these findings demonstrate the efficacy of AIO-P in the NAS setting for any general neural network, not simply for well-known search spaces.

\section{Conclusion}
\label{sec:conclusion}

We propose AIO-P, or All-in-One Predictors, to broaden the scope of neural performance prediction tasks. AIO-P uses $K$-Adapters to infuse knowledge from different tasks and accepts Computational Graphs (CG) as input. CGs represent the body and head of an architecture by encoding all atomic operations as nodes with directed edges determined by the network forward pass. At the output, AIO-P incorporates target scaling techniques, including one based on FLOPs, to re-scale predictions into the appropriate task metric and ultimately obtain superior performance. To construct a suitable training set, we devise a shared head approach with latent sampling, which can pair with any architecture in the search space to produce a pseudo-label that is highly correlated with the true label. Experimental results show that AIO-P can obtain Mean Absolute Error and Spearman's Rank Correlations below 1\% and above 0.5, respectively, when transferred to a wide range of downstream tasks. Moreover, AIO-P can directly transfer and adequately rank different networks in several foreign model zoos not seen in training for classification and semantic segmentation. Finally, we use AIO-P to optimize a proprietary facial recognition network to effectively preserve precision and recall while reducing FLOPs by over 13.5\%.

\begin{table}[t]
    \centering
    \scalebox{0.8}{
    \begin{tabular}{l|cc|cc|cc|c} \toprule
     & \textbf{Full} & \textbf{Simple} & \textbf{Lighted} & \textbf{Dark} & \textbf{FLOPs}\\ \midrule
    Base Model Pr & \textbf{96.3}\% & \textbf{98.7}\% & \textbf{97.9}\% & 96.5\% & 563M \\
    AIO-P Search Pr & 96.1\% & \textbf{98.7}\% & \textbf{97.9}\% & \textbf{96.7}\% & 486M\\ \midrule
    Base Model Rc & \textbf{91.9}\% & \textbf{98.3}\% & \textbf{96.8}\% & 92.6\% & 563M \\
    AIO-P Search Rc & 91.1\% & 98.2\% & 96.6\% & \textbf{93.2}\% & 486M\\ \bottomrule
    \end{tabular}
    }
    \caption{Precision (Pr) and Recall (Rc) of a proprietary FR network found by pairing AIO-P with a search algorithm designed to preserve performance while reducing FLOPs.}
    \label{tab:properitary}
\end{table}

\bibliography{references}

\begin{thebibliography}{54}
\providecommand{\natexlab}[1]{#1}

\bibitem[{Abadi et~al.(2015)Abadi, Agarwal, Barham, Brevdo, Chen, Citro,
  Corrado, Davis, Dean, Devin, Ghemawat, Goodfellow, Harp, Irving, Isard, Jia,
  Jozefowicz, Kaiser, Kudlur, Levenberg, Man\'{e}, Monga, Moore, Murray, Olah,
  Schuster, Shlens, Steiner, Sutskever, Talwar, Tucker, Vanhoucke, Vasudevan,
  Vi\'{e}gas, Vinyals, Warden, Wattenberg, Wicke, Yu, and
  Zheng}]{tensorflow2015-whitepaper}
Abadi, M.; Agarwal, A.; Barham, P.; Brevdo, E.; Chen, Z.; Citro, C.; Corrado,
  G.~S.; Davis, A.; Dean, J.; Devin, M.; Ghemawat, S.; Goodfellow, I.; Harp,
  A.; Irving, G.; Isard, M.; Jia, Y.; Jozefowicz, R.; Kaiser, L.; Kudlur, M.;
  Levenberg, J.; Man\'{e}, D.; Monga, R.; Moore, S.; Murray, D.; Olah, C.;
  Schuster, M.; Shlens, J.; Steiner, B.; Sutskever, I.; Talwar, K.; Tucker, P.;
  Vanhoucke, V.; Vasudevan, V.; Vi\'{e}gas, F.; Vinyals, O.; Warden, P.;
  Wattenberg, M.; Wicke, M.; Yu, Y.; and Zheng, X. 2015.
\newblock {TensorFlow}: Large-Scale Machine Learning on Heterogeneous Systems.

\bibitem[{Abdelfattah et~al.(2021)Abdelfattah, Mehrotra, Dudziak, and
  Lane}]{abdelfattah2021zero}
Abdelfattah, M.~S.; Mehrotra, A.; Dudziak, {\L}.; and Lane, N.~D. 2021.
\newblock {Zero-Cost Proxies for Lightweight NAS}.
\newblock In \emph{International Conference on Learning Representations
  (ICLR)}.

\bibitem[{Andriluka et~al.(2014)Andriluka, Pishchulin, Gehler, and
  Schiele}]{MPII}
Andriluka, M.; Pishchulin, L.; Gehler, P.; and Schiele, B. 2014.
\newblock 2D Human Pose Estimation: New Benchmark and State of the Art
  Analysis.
\newblock In \emph{IEEE Conference on Computer Vision and Pattern Recognition
  (CVPR)}.

\bibitem[{Artacho and Savakis(2020)}]{artacho2020unipose}
Artacho, B.; and Savakis, A. 2020.
\newblock UniPose: Unified Human Pose Estimation in Single Images and Videos.
\newblock In \emph{Proceedings of the IEEE/CVF Conference on Computer Vision
  and Pattern Recognition}, 7035--7044.

\bibitem[{Cai et~al.(2020)Cai, Gan, Wang, Zhang, and Han}]{cai2020once}
Cai, H.; Gan, C.; Wang, T.; Zhang, Z.; and Han, S. 2020.
\newblock Once for All: Train One Network and Specialize it for Efficient
  Deployment.
\newblock In \emph{International Conference on Learning Representations}.

\bibitem[{Cai, Zhu, and Han(2019)}]{cai2018proxylessnas}
Cai, H.; Zhu, L.; and Han, S. 2019.
\newblock Proxyless{NAS}: Direct Neural Architecture Search on Target Task and
  Hardware.
\newblock In \emph{International Conference on Learning Representations}.

\bibitem[{Changiz~Rezaei et~al.(2021)Changiz~Rezaei, Han, Niu, Salameh, Mills,
  Lian, Lu, and Jui}]{rezaei2021generative}
Changiz~Rezaei, S.~S.; Han, F.~X.; Niu, D.; Salameh, M.; Mills, K.; Lian, S.;
  Lu, W.; and Jui, S. 2021.
\newblock Generative Adversarial Neural Architecture Search.
\newblock In Zhou, Z.-H., ed., \emph{Proceedings of the Thirtieth International
  Joint Conference on Artificial Intelligence, {IJCAI-21}}, 2227--2234.
  International Joint Conferences on Artificial Intelligence Organization.
\newblock Main Track.

\bibitem[{Chen et~al.(2018)Chen, Zhu, Papandreou, Schroff, and
  Adam}]{deeplabv3plus2018}
Chen, L.-C.; Zhu, Y.; Papandreou, G.; Schroff, F.; and Adam, H. 2018.
\newblock Encoder-Decoder With Atrous Separable Convolution For Semantic Image
  Segmentation.
\newblock In \emph{Proceedings of the European Conference on Computer Vision
  (ECCV)}, 801--818.

\bibitem[{Chollet(2017)}]{chollet2017xception}
Chollet, F. 2017.
\newblock Xception: Deep Learning with Depthwise Separable Convolutions.
\newblock In \emph{Proceedings of the IEEE Conference on Computer Vision and
  Pattern Recognition}, 1251--1258.

\bibitem[{Cordts et~al.(2016)Cordts, Omran, Ramos, Rehfeld, Enzweiler,
  Benenson, Franke, Roth, and Schiele}]{Cordts2016Cityscapes}
Cordts, M.; Omran, M.; Ramos, S.; Rehfeld, T.; Enzweiler, M.; Benenson, R.;
  Franke, U.; Roth, S.; and Schiele, B. 2016.
\newblock The Cityscapes Dataset for Semantic Urban Scene Understanding.
\newblock In \emph{Proc. of the IEEE Conference on Computer Vision and Pattern
  Recognition (CVPR)}.

\bibitem[{Ding et~al.(2022)Ding, Huo, Lu, Yang, Wang, Lu, Wang, and
  Luo}]{ding2021learning}
Ding, M.; Huo, Y.; Lu, H.; Yang, L.; Wang, Z.; Lu, Z.; Wang, J.; and Luo, P.
  2022.
\newblock Learning Versatile Neural Architectures by Propagating Network Codes.
\newblock In \emph{International Conference on Learning Representations}.

\bibitem[{Dong and Yang(2020)}]{dong2020nasbench201}
Dong, X.; and Yang, Y. 2020.
\newblock NAS-Bench-201: Extending the Scope of Reproducible Neural
  Architecture Search.
\newblock In \emph{International Conference on Learning Representations}.

\bibitem[{Duan et~al.(2021)Duan, Chen, Xu, Chen, Liang, Zhang, and
  Li}]{duan2021transnas}
Duan, Y.; Chen, X.; Xu, H.; Chen, Z.; Liang, X.; Zhang, T.; and Li, Z. 2021.
\newblock TransNAS-Bench-101: Improving Transferability and Generalizability of
  Cross-Task Neural Architecture Search.
\newblock In \emph{Proceedings of the IEEE/CVF Conference on Computer Vision
  and Pattern Recognition}, 5251--5260.

\bibitem[{Elsken et~al.(2019)Elsken, Metzen, Hutter et~al.}]{elsken2019neural}
Elsken, T.; Metzen, J.~H.; Hutter, F.; et~al. 2019.
\newblock Neural Architecture Search: A Survey.
\newblock \emph{J. Mach. Learn. Res.}, 20(55): 1--21.

\bibitem[{Everingham et~al.(2015)Everingham, Eslami, Van~Gool, Williams, Winn,
  and Zisserman}]{everingham2015Pascal}
Everingham, M.; Eslami, S. M.~A.; Van~Gool, L.; Williams, C. K.~I.; Winn, J.;
  and Zisserman, A. 2015.
\newblock The Pascal Visual Object Classes Challenge: A Retrospective.
\newblock \emph{International Journal of Computer Vision}, 111(1): 98--136.

\bibitem[{He et~al.(2017)He, Gkioxari, Doll{\'a}r, and Girshick}]{he2017mask}
He, K.; Gkioxari, G.; Doll{\'a}r, P.; and Girshick, R. 2017.
\newblock Mask R-CNN.
\newblock In \emph{Proceedings of the IEEE International Conference on Computer
  Vision}, 2961--2969.

\bibitem[{He et~al.(2016{\natexlab{a}})He, Zhang, Ren, and Sun}]{he2016deep}
He, K.; Zhang, X.; Ren, S.; and Sun, J. 2016{\natexlab{a}}.
\newblock Deep Residual Learning for Image Recognition.
\newblock In \emph{Proceedings of the IEEE Conference on Computer Vision and
  Pattern Recognition}, 770--778.

\bibitem[{He et~al.(2016{\natexlab{b}})He, Zhang, Ren, and
  Sun}]{he2016identity}
He, K.; Zhang, X.; Ren, S.; and Sun, J. 2016{\natexlab{b}}.
\newblock Identity Mappings in Deep Residual Networks.
\newblock In \emph{European Conference on Computer Vision}, 630--645. Springer.

\bibitem[{Howard et~al.(2019)Howard, Sandler, Chu, Chen, Chen, Tan, Wang, Zhu,
  Pang, Vasudevan et~al.}]{howard2019searching}
Howard, A.; Sandler, M.; Chu, G.; Chen, L.-C.; Chen, B.; Tan, M.; Wang, W.;
  Zhu, Y.; Pang, R.; Vasudevan, V.; et~al. 2019.
\newblock Searching For MobileNetV3.
\newblock In \emph{Proceedings of the IEEE/CVF International Conference on
  Computer Vision}, 1314--1324.

\bibitem[{Hu, Shen, and Sun(2018)}]{hu2018squeeze}
Hu, J.; Shen, L.; and Sun, G. 2018.
\newblock Squeeze-and-Excitation Networks.
\newblock In \emph{Proceedings of the IEEE Conference on Computer Vision and
  Pattern Recognition}, 7132--7141.

\bibitem[{Johnson and Everingham(2011)}]{lsp_extended}
Johnson, S.; and Everingham, M. 2011.
\newblock Learning Effective Human Pose Estimation from Inaccurate Annotation.
\newblock In \emph{IEEE Conference on Computer Vision and Pattern Recognition
  (CVPR)}, 1465--1472. IEEE.

\bibitem[{Kirillov et~al.(2019)Kirillov, Girshick, He, and
  Doll{\'a}r}]{kirillov2019panoptic}
Kirillov, A.; Girshick, R.; He, K.; and Doll{\'a}r, P. 2019.
\newblock Panoptic Feature Pyramid Networks.
\newblock In \emph{Proceedings of the IEEE/CVF Conference on Computer Vision
  and Pattern Recognition}, 6399--6408.

\bibitem[{Krizhevsky, Hinton et~al.(2009)}]{Krizhevsky09CIFAR}
Krizhevsky, A.; Hinton, G.; et~al. 2009.
\newblock Learning Multiple Layers of Features From Tiny Images.
\newblock \emph{Technical Report}.

\bibitem[{Lin et~al.(2017)Lin, Doll{\'a}r, Girshick, He, Hariharan, and
  Belongie}]{lin2017feature}
Lin, T.-Y.; Doll{\'a}r, P.; Girshick, R.; He, K.; Hariharan, B.; and Belongie,
  S. 2017.
\newblock Feature Pyramid Networks for Object Detection.
\newblock In \emph{Proceedings of the IEEE Conference on Computer Vision and
  Pattern Recognition}, 2117--2125.

\bibitem[{Lin et~al.(2014)Lin, Maire, Belongie, Bourdev, Girshick, Hays,
  Perona, Ramanan, Zitnick, and Doll{\'a}r}]{coco}
Lin, T.-Y.; Maire, M.; Belongie, S.; Bourdev, L.; Girshick, R.; Hays, J.;
  Perona, P.; Ramanan, D.; Zitnick, C.~L.; and Doll{\'a}r, P. 2014.
\newblock Microsoft COCO: Common Objects in Context.
\newblock In \emph{Computer Vision -- ECCV 2014}, 740--755. Cham: Springer
  International Publishing.

\bibitem[{Liu et~al.(2019)Liu, Chen, Schroff, Adam, Hua, Yuille, and
  Fei-Fei}]{liu2019auto}
Liu, C.; Chen, L.-C.; Schroff, F.; Adam, H.; Hua, W.; Yuille, A.~L.; and
  Fei-Fei, L. 2019.
\newblock Auto-DeepLab: Hierarchical Neural Architecture Search for Semantic
  Image Segmentation.
\newblock In \emph{Proceedings of the IEEE/CVF Conference on Computer Vision
  and Pattern Recognition}, 82--92.

\bibitem[{Liu, Simonyan, and Yang(2019)}]{liu2018DARTS}
Liu, H.; Simonyan, K.; and Yang, Y. 2019.
\newblock DARTS: Differentiable Architecture Search.
\newblock In \emph{International Conference on Learning Representations
  (ICLR)}.

\bibitem[{Loshchilov and Hutter(2017)}]{Loshchilov2017SGDRSG}
Loshchilov, I.; and Hutter, F. 2017.
\newblock {SGDR:} Stochastic Gradient Descent with Warm Restarts.
\newblock In \emph{5th International Conference on Learning Representations,
  {ICLR} 2017, Toulon, France, April 24-26, 2017, Conference Track
  Proceedings}.

\bibitem[{Lu et~al.(2021{\natexlab{a}})Lu, Yang, Jiang, Shi, and
  Ren}]{lu2021one}
Lu, B.; Yang, J.; Jiang, W.; Shi, Y.; and Ren, S. 2021{\natexlab{a}}.
\newblock One Proxy Device is Enough for Hardware-Aware Neural Architecture
  Search.
\newblock \emph{Proceedings of the ACM on Measurement and Analysis of Computing
  Systems}, 5(3): 1--34.

\bibitem[{Lu et~al.(2021{\natexlab{b}})Lu, Li, Tan, Yang, and
  Liu}]{lu2021TNASP}
Lu, S.; Li, J.; Tan, J.; Yang, S.; and Liu, J. 2021{\natexlab{b}}.
\newblock TNASP: A Transformer-based NAS Predictor with a Self-Evolution
  Framework.
\newblock In Ranzato, M.; Beygelzimer, A.; Dauphin, Y.; Liang, P.; and Vaughan,
  J.~W., eds., \emph{Advances in Neural Information Processing Systems},
  volume~34, 15125--15137. Curran Associates, Inc.

\bibitem[{Luo et~al.(2020)Luo, Tan, Wang, Qin, Chen, and Liu}]{luo2020semi}
Luo, R.; Tan, X.; Wang, R.; Qin, T.; Chen, E.; and Liu, T.-Y. 2020.
\newblock Semi-Supervised Neural Architecture Search.
\newblock \emph{Advances in Neural Information Processing Systems}, 33:
  10547--10557.

\bibitem[{Mills et~al.(2021{\natexlab{a}})Mills, Han, Salameh, Changiz~Rezaei,
  Kong, Lu, Lian, Jui, and Niu}]{mills2021l2nas}
Mills, K.~G.; Han, F.~X.; Salameh, M.; Changiz~Rezaei, S.~S.; Kong, L.; Lu, W.;
  Lian, S.; Jui, S.; and Niu, D. 2021{\natexlab{a}}.
\newblock L$^2$NAS: Learning to Optimize Neural Architectures via
  Continuous-Action Reinforcement Learning.
\newblock In \emph{Proceedings of the 30th ACM International Conference on
  Information \& Knowledge Management}, 1284--1293.

\bibitem[{Mills et~al.(2021{\natexlab{b}})Mills, Han, Zhang, Changiz~Rezaei,
  Chudak, Lu, Lian, Jui, and Niu}]{mills2021profiling}
Mills, K.~G.; Han, F.~X.; Zhang, J.; Changiz~Rezaei, S.~S.; Chudak, F.; Lu, W.;
  Lian, S.; Jui, S.; and Niu, D. 2021{\natexlab{b}}.
\newblock Profiling Neural Blocks and Design Spaces for Mobile Neural
  Architecture Search.
\newblock In \emph{Proceedings of the 30th ACM International Conference on
  Information \& Knowledge Management}, 4026--4035.

\bibitem[{Mills et~al.(2021{\natexlab{c}})Mills, Salameh, Niu, Han, Rezaei,
  Yao, Lu, Lian, and Jui}]{mills2021exploring}
Mills, K.~G.; Salameh, M.; Niu, D.; Han, F.~X.; Rezaei, S. S.~C.; Yao, H.; Lu,
  W.; Lian, S.; and Jui, S. 2021{\natexlab{c}}.
\newblock Exploring Neural Architecture Search Space via Deep Deterministic
  Sampling.
\newblock \emph{IEEE Access}, 9: 110962--110974.

\bibitem[{Morris et~al.(2019)Morris, Ritzert, Fey, Hamilton, Lenssen, Rattan,
  and Grohe}]{morris2019weisfeiler}
Morris, C.; Ritzert, M.; Fey, M.; Hamilton, W.~L.; Lenssen, J.~E.; Rattan, G.;
  and Grohe, M. 2019.
\newblock Weisfeiler and Leman Go Neural: Higher-Order Graph Neural Networks.
\newblock In \emph{Proceedings of the AAAI Conference on Artificial
  Intelligence}, volume~33, 4602--4609.

\bibitem[{Munoz et~al.(2022)Munoz, Lyalyushkin, Lacewell, Senina, Cummings,
  Sarah, Kozlov, and Jain}]{munoz2022automated}
Munoz, J.~P.; Lyalyushkin, N.; Lacewell, C.~W.; Senina, A.; Cummings, D.;
  Sarah, A.; Kozlov, A.; and Jain, N. 2022.
\newblock Automated Super-Network Generation for Scalable Neural Architecture
  Search.
\newblock In \emph{International Conference on Automated Machine Learning},
  5--1. PMLR.

\bibitem[{Pham et~al.(2018)Pham, Guan, Zoph, Le, and Dean}]{pham2018ENAS}
Pham, H.; Guan, M.; Zoph, B.; Le, Q.; and Dean, J. 2018.
\newblock Efficient Neural Architecture Search via Parameters Sharing.
\newblock In \emph{International Conference on Machine Learning}, 4095--4104.
  PMLR.

\bibitem[{Ren et~al.(2015)Ren, He, Girshick, and Sun}]{ren2015faster}
Ren, S.; He, K.; Girshick, R.; and Sun, J. 2015.
\newblock Faster R-CNN: Towards Real-Time Object Detection with Region Proposal
  Networks.
\newblock \emph{Advances in Neural Information Processing Systems}, 28.

\bibitem[{Russakovsky et~al.(2015)Russakovsky, Deng, Su, Krause, Satheesh, Ma,
  Huang, Karpathy, Khosla, Bernstein et~al.}]{russakovsky2015imagenet}
Russakovsky, O.; Deng, J.; Su, H.; Krause, J.; Satheesh, S.; Ma, S.; Huang, Z.;
  Karpathy, A.; Khosla, A.; Bernstein, M.; et~al. 2015.
\newblock ImageNet Large Scale Visual Recognition Challenge.
\newblock \emph{International Journal of Computer Vision}, 115(3): 211--252.

\bibitem[{{Sergio Guadarrama, Nathan Silberman}(2016)}]{TFSlim}
{Sergio Guadarrama, Nathan Silberman}. 2016.
\newblock {TensorFlow-Slim}: A Lightweight Library for Defining, Training and
  Evaluating Complex Models in TensorFlow.

\bibitem[{Szegedy et~al.(2017)Szegedy, Ioffe, Vanhoucke, and
  Alemi}]{szegedy2017inception}
Szegedy, C.; Ioffe, S.; Vanhoucke, V.; and Alemi, A.~A. 2017.
\newblock Inception-V4, Inception-ResNet and the Impact of Residual Connections
  on Learning.
\newblock In \emph{Thirty-first AAAI Conference on Artificial Intelligence}.

\bibitem[{Tan and Le(2019)}]{tan2019efficientnet}
Tan, M.; and Le, Q. 2019.
\newblock EfficientNet: Rethinking Model Scaling for Convolutional Neural
  Networks.
\newblock In \emph{International Conference on Machine Learning}, 6105--6114.
  PMLR.

\bibitem[{Tanaka et~al.(2020)Tanaka, Kunin, Yamins, and
  Ganguli}]{tanaka2020pruning}
Tanaka, H.; Kunin, D.; Yamins, D.~L.; and Ganguli, S. 2020.
\newblock Pruning Neural Networks Without Any Data by Iteratively Conserving
  Synaptic Flow.
\newblock \emph{Advances in Neural Information Processing Systems}, 33:
  6377--6389.

\bibitem[{Wang et~al.(2021)Wang, Tang, Duan, Wei, Huang, Ji, Cao, Jiang, and
  Zhou}]{wang2021kadapt}
Wang, R.; Tang, D.; Duan, N.; Wei, Z.; Huang, X.; Ji, J.; Cao, G.; Jiang, D.;
  and Zhou, M. 2021.
\newblock {K-Adapter}: {I}nfusing {K}nowledge into {P}re-{T}rained {M}odels
  with {A}dapters.
\newblock In \emph{Findings of the Association for Computational Linguistics:
  ACL-IJCNLP 2021}, 1405--1418. Online: Association for Computational
  Linguistics.

\bibitem[{Wen et~al.(2020)Wen, Liu, Chen, Li, Bender, and
  Kindermans}]{wen2020neural}
Wen, W.; Liu, H.; Chen, Y.; Li, H.; Bender, G.; and Kindermans, P.-J. 2020.
\newblock Neural Predictor for Neural Architecture Search.
\newblock In \emph{European Conference on Computer Vision}, 660--676. Springer.

\bibitem[{White, Neiswanger, and Savani(2021)}]{white2019bananas}
White, C.; Neiswanger, W.; and Savani, Y. 2021.
\newblock BANANAS: Bayesian Optimization with Neural Architectures for Neural
  Architecture Search.
\newblock In \emph{Proceedings of the AAAI Conference on Artificial
  Intelligence}, volume~35, 10293--10301.

\bibitem[{Wu et~al.(2021)Wu, Dai, Chen, Chen, Liu, Yu, Wang, Liu, Chen, and
  Yuan}]{wu2021stronger}
Wu, J.; Dai, X.; Chen, D.; Chen, Y.; Liu, M.; Yu, Y.; Wang, Z.; Liu, Z.; Chen,
  M.; and Yuan, L. 2021.
\newblock Stronger NAS with Weaker Predictors.
\newblock \emph{Advances in Neural Information Processing Systems}, 34:
  28904--28918.

\bibitem[{Wu et~al.(2019)Wu, Kirillov, Massa, Lo, and
  Girshick}]{wu2019detectron2}
Wu, Y.; Kirillov, A.; Massa, F.; Lo, W.-Y.; and Girshick, R. 2019.
\newblock Detectron2.
\newblock \url{https://github.com/facebookresearch/detectron2}.

\bibitem[{Ying et~al.(2019)Ying, Klein, Christiansen, Real, Murphy, and
  Hutter}]{ying2019nasbench101}
Ying, C.; Klein, A.; Christiansen, E.; Real, E.; Murphy, K.; and Hutter, F.
  2019.
\newblock NAS-Bench-101: Towards Reproducible Neural Architecture Search.
\newblock In \emph{International Conference on Machine Learning}, 7105--7114.

\bibitem[{Zela et~al.(2022)Zela, Siems, Zimmer, Lukasik, Keuper, and
  Hutter}]{zela2022surrogate}
Zela, A.; Siems, J.~N.; Zimmer, L.; Lukasik, J.; Keuper, M.; and Hutter, F.
  2022.
\newblock Surrogate {NAS} Benchmarks: Going Beyond the Limited Search Spaces of
  Tabular {NAS} Benchmarks.
\newblock In \emph{International Conference on Learning Representations}.

\bibitem[{Zheng et~al.(2020)Zheng, Wu, Chen, Yang, Zhu, Shen, Kehtarnavaz, and
  Shah}]{zheng2020deep}
Zheng, C.; Wu, W.; Chen, C.; Yang, T.; Zhu, S.; Shen, J.; Kehtarnavaz, N.; and
  Shah, M. 2020.
\newblock Deep Learning-Based Human Pose Estimation: A Survey.

\bibitem[{Zhou et~al.(2017{\natexlab{a}})Zhou, Zhao, Puig, Fidler, Barriuso,
  and Torralba}]{zhou2017ADE}
Zhou, B.; Zhao, H.; Puig, X.; Fidler, S.; Barriuso, A.; and Torralba, A.
  2017{\natexlab{a}}.
\newblock Scene Parsing Through ADE20K Dataset.
\newblock In \emph{2017 IEEE Conference on Computer Vision and Pattern
  Recognition (CVPR)}, 5122--5130.

\bibitem[{Zhou et~al.(2017{\natexlab{b}})Zhou, Huang, Sun, Xue, and
  Wei}]{Zhou_2017_ICCV}
Zhou, X.; Huang, Q.; Sun, X.; Xue, X.; and Wei, Y. 2017{\natexlab{b}}.
\newblock Towards 3D Human Pose Estimation in the Wild: A Weakly-Supervised
  Approach.
\newblock In \emph{The IEEE International Conference on Computer Vision
  (ICCV)}.

\bibitem[{Zoph and Le(2017)}]{zoph2017NAS}
Zoph, B.; and Le, Q. 2017.
\newblock Neural Architecture Search with Reinforcement Learning.
\newblock In \emph{International Conference on Learning Representations}.

\end{thebibliography}

\clearpage
\newpage
\section{Supplementary Material}
\label{sec:supp}

\subsection{Summary of Search Spaces and Tasks}

\begin{figure}[H]
    \centering
    \subfloat[]{\includegraphics[width=3.2in]{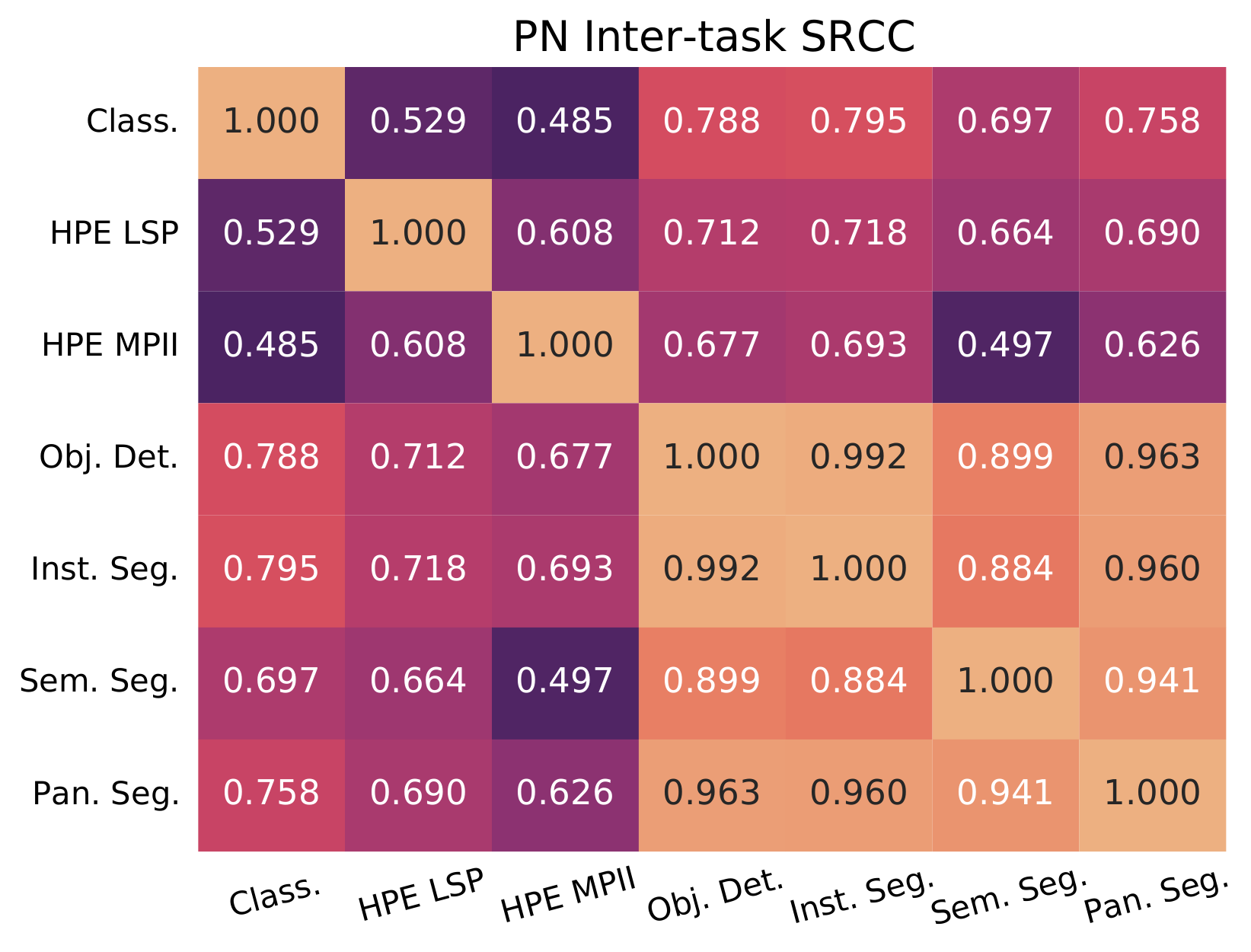}}\hspace{0mm}
    \subfloat[]{\includegraphics[width=3.2in]{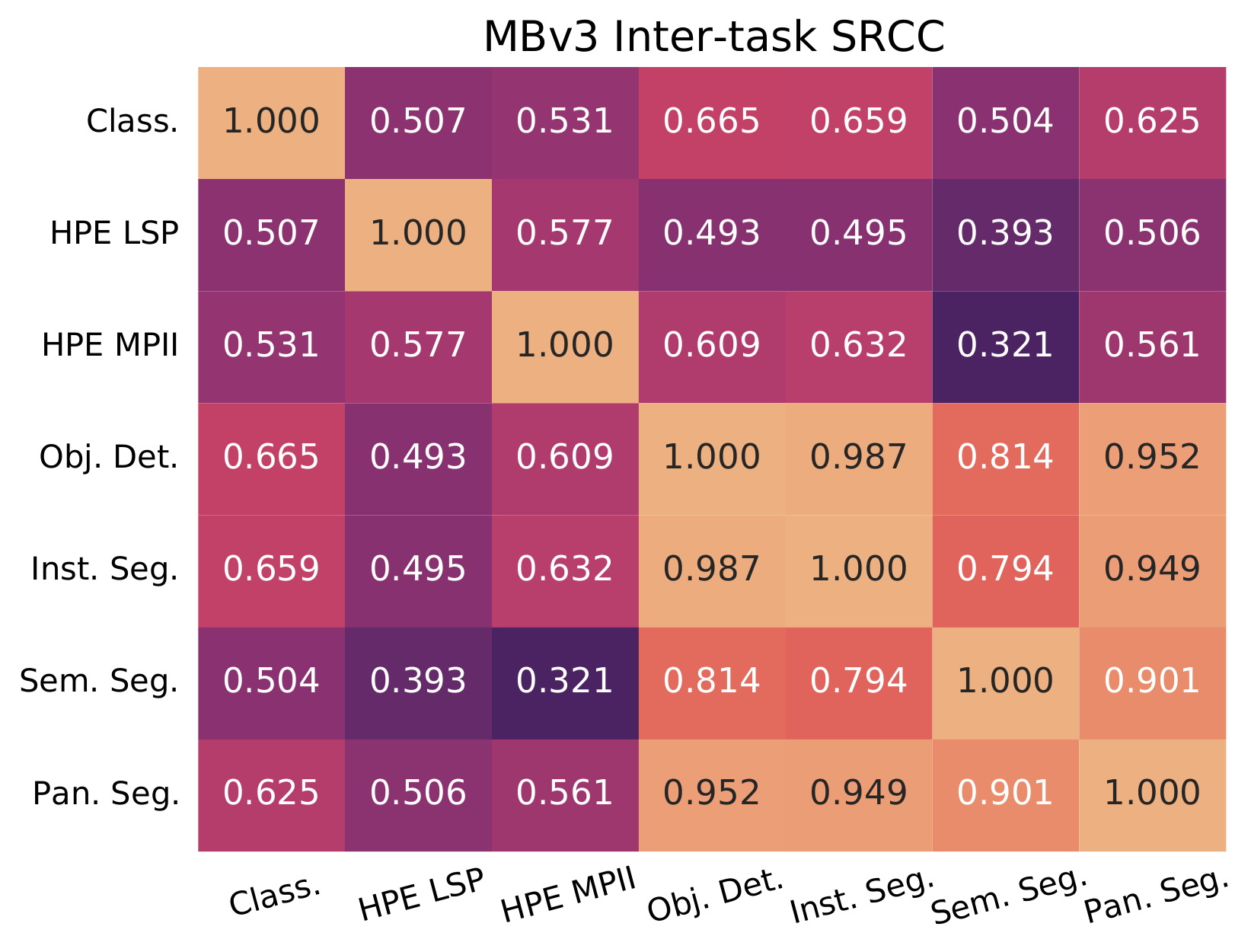}}\hspace{0mm}
    \subfloat[]{\includegraphics[width=3.2in]{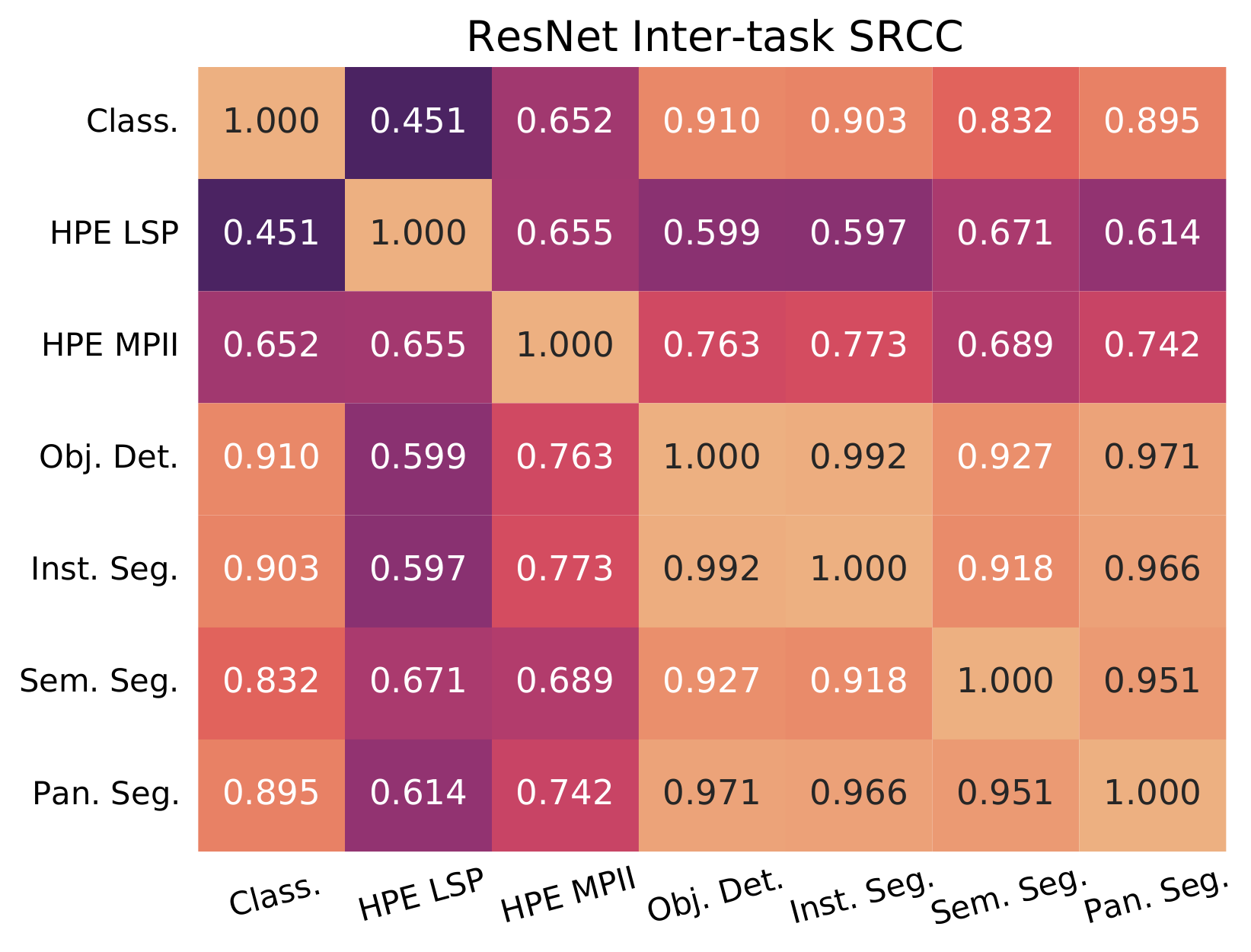}}
    \caption{Inter-task correlations for ProxylessNAS (a), MobileNetV3 (b) and ResNet50 (c). We measure SRCC by comparing the performance values of individually trained architectures across all tasks.}
    \label{fig:intertask_srcc}
\end{figure}

Figure~\ref{fig:intertask_srcc} illustrates SRCC between different tasks on each search space. Table~\ref{tab:app_full_architecture_description} enumerates the number of architectures we individually train across all search spaces and tasks. It also provides distributions for task performance, FLOPs, and parameters. Moreover, Table~\ref{tab:app_fps} reports Frames Per Second (FPS) distributions all search spaces performing Panoptic Segmentation. Finally, Table~\ref{tab:app_shared_head_architecture_description} provides a statistical breakdown 
for architectures fine-tuned using latent sampling shared heads.

\subsection{Training Architectures Individually}
We form test sets for AIO-P by pairing bodies with task heads and training them individually following hyperparameters and learning schedules from previous works. 

\subsubsection{Detectron2 (OD/IS/SS/PS)}

Using Detectron2~\cite{wu2019detectron2}, we can train architectures to perform Object Detection (OD), Instance Segmentation, Semantic Segmentation (SS) and Panoptic Segmentation (PS) all at once. Specifically, we adopt the training regime described by the configuration files \texttt{panoptic\_fpn\_R\_50\_1x.yaml}, \texttt{Base-Panoptic-FPN.yaml} as well as \texttt{Base-RCNN-FPN.yaml} when training OFA architectures. This setup includes all four input levels of the Feature Pyramid Net (FPN)~\cite{lin2017feature}, `$P^2$' through `$P^5$' (see the `Task Head Descriptions' subsection for a definition of these terms). Our FPN implementation performs additional downsampling to generate `$P^6$' from `$P^5$'. Specifically, this setup involves freezing pre-trained weights of the first two `stages' of the network that use the highest resolution feature tensors. Under this configuration, an architecture trains using a batch size of 16 for 90k steps with a base learning rate of 0.02, which we reduce at 60k and 80k by a factor of 10. It takes between 24 and 30 hours to train and evaluate a single architecture using 2 Tesla V100 GPUs with 32GB of VRAM, and ResNets take longer than MobileNets.

\subsubsection{Human Pose Estimation (HPE)}

We adopt the default parameters of \cite{Zhou_2017_ICCV} to perform 2D Human Pose Estimation (HPE) on MPII~\cite{MPII} and Leeds Sports Pose-Extended (LSP)~\cite{lsp_extended}. Specifically, we train an architecture for 140 epochs on a given dataset with a batch size of 32. The initial learning rate is 0.001, which we reduce by a factor of 10 at epochs 90 and 120. The size of an input image is 256 $\times$ 256 in this setting, and bodies from all OFA search spaces will downsample it down to a resolution of 8 $\times$ 8, after which we use a series of deconvolutions to upsample and produce joint heatmaps of size 64 $\times$ 64. MPII contains $J = 16$ joints by default, while LSP only contains 14. We adopt a technique from \cite{artacho2020unipose} to estimate labels for the thorax and hip, the two remaining joints. Training an architecture takes a few hours under these settings. 

\begin{table*}
    \centering
    \caption{Summary of statistics for individually trained architectures across all three OFA families and CV tasks in terms of number of architectures, task performance [\%], and resource metrics such as FLOPs [G] and parameters [M] for all tasks. For all metrics, we report the mean and standard deviation on one line, and the range on the second line.}
    \label{tab:app_full_architecture_description}
    \scalebox{0.7}{
    \begin{tabular}{l|c|c|c|c|c|c|c} \toprule
    \textbf{Family} & \textbf{Classification} & \textbf{LSP HPE} & \textbf{MPII HPE} & \textbf{Obj. Det.} & \textbf{Inst. Seg.} & \textbf{Sem. Seg.} & \textbf{Pan. Seg.} \\ \midrule \midrule
    PN (\# Archs) & 8.2k & 215 & 246 & 118 & 118 & 118 & 118 \\ \midrule
    Performance [\%] & 75.41  $\pm$  0.09 & 65.16  $\pm$  0.80 & 84.33  $\pm$  0.47 & 33.37  $\pm$  1.01 & 31.04  $\pm$  0.82 & 39.61  $\pm$  0.72 & 36.90  $\pm$  0.73 \\
    & [71.15, 77.81] &[62.77, 66.92] &[82.79, 85.15] &[30.17, 35.62] &[28.33, 32.85] &[37.10, 40.85] &[34.25, 38.49] \\ \midrule
    FLOPs [G]        &  1.12 $\pm$ 0.14  & 15.73  $\pm$  0.17 & 15.73  $\pm$  0.17 & 86.82  $\pm$  1.07 & 125.50  $\pm$  1.07 & 63.64  $\pm$  1.07 & 174.02  $\pm$  1.07 \\
    & [0.69, 1.59] & [15.28, 16.13] &[15.28, 16.26] &[84.09, 89.25] &[122.77, 127.93] &[60.91, 66.07] &[171.29, 176.45] \\ \midrule
    Params [M]       &  5.30 $\pm$ 0.60  & 12.54  $\pm$  0.60 & 12.55  $\pm$  0.61 & 32.36  $\pm$  0.56 & 20.70  $\pm$  0.56 & 20.13  $\pm$  0.56 & 37.67  $\pm$ 0.56 \\
    & [3.88, 7.18] & [11.32, 14.13] &[11.32, 14.13] &[31.43, 33.52] &[19.76, 21.86] &[19.19, 21.29] &[36.73, 38.83] \\ \midrule
    \midrule
    MBv3 (\# Archs) & 7.5k & 217 & 236 & 118 & 118 & 118 & 118 \\ \midrule
    Performance [\%] & 76.94  $\pm$  0.08 & 65.22  $\pm$  0.77 & 83.38 $\pm$  0.46& 33.39 $\pm$  0.89& 31.25 $\pm$  0.70& 39.27  $\pm$  0.90& 36.88 $\pm$ 0.73 \\
    & [73.56, 78.83] &[62.27, 67.14] &[81.82, 84.39] &[30.54, 35.38] &[28.93, 32.87] &[36.94, 41.07] &[35.02, 38.50] \\ \midrule
    FLOPs [G]        &  0.90 $\pm$ 0.13 & 14.36 $\pm$ 0.16 & 14.36 $\pm$ 0.16 & 84.70 $\pm$ 1.00 & 123.38 $\pm$  1.00 & 61.52 $\pm$  1.00 & 171.84 $\pm$ 1.00 \\
    & [0.52, 1.37] &[13.95, 14.76] &[13.95, 14.76] &[82.06, 87.20] &[120.74, 125.89] &[58.88, 64.02] & [169.19, 174.34] \\ \midrule
    Params [M]       & 6.71 $\pm$ 0.85 & 10.26 $\pm$ 0.88 & 10.24 $\pm$ 0.88 & 36.48 $\pm$ 0.85 & 24.82 $\pm$ 0.85 & 24.24 $\pm$ 0.85 & 41.79 $\pm$ 0.85 \\
    & [4.71, 9.89] &[8.51, 12.43] &[8.51, 12.43] &[34.73, 38.64] &[23.06, 26.98] &[22.49, 26.41] &[58.88, 64.02] \\ \midrule
    \midrule
    R50 (\# Archs) & 10k & 215 & 236 & 115 & 115 & 115 & 115 \\ \midrule
    Performance [\%] & 78.19  $\pm$  0.07 & 65.64 $\pm$ 0.88 & 85.40 $\pm$ 0.40 & 35.86 $\pm$ 1.02 & 32.88 $\pm$ 0.80 & 39.41 $\pm$ 0.71 & 37.98 $\pm$ 0.71 \\
    & [75.25, 79.94] &[62.67, 67.80] &[84.14, 86.28] &[32.97, 38.18] &[30.58, 34.57] &[37.58, 40.84] &[35.81, 39.49] \\ \midrule
    FLOPs [G]        & 5.89 $\pm$ 1.22 & 22.17 $\pm$ 1.97 & 22.16 $\pm$ 1.99 & 133.94 $\pm$ 11.39 & 172.62 $\pm$ 11.39 & 110.76 $\pm$ 11.39 & 221.14 $\pm$ 11.39 \\
    & [2.67, 11.39] &[17.68, 27.60] &[17.68, 27.60] & [110.88, 164.95] & [149.57, 203.64] &[87.70, 141.77] & [198.08, 252.15] \\ \midrule
    Params [M]       & 18.68 $\pm$ 5.79 & 26.57 $\pm$ 6.48 &26.63  $\pm$ 6.60 & 86.73 $\pm$ 6.07 & 75.06 $\pm$ 6.07 & 74.49 $\pm$ 6.07 & 92.03 $\pm$ 6.07 \\
    & [7.19, 41.15] &[14.84, 47.75] &[14.84, 47.76] &[76.34, 107.00] &[64.67, 95.34] &[64.10, 94.76] &[81.64, 141.77] \\ 
    \bottomrule
    \end{tabular}
    }
\end{table*}

\begin{table}
    \centering
    \caption{Average Frames Per Second for Panoptic Segmentation. We measure on an NVIDIA V100 GPU with 32GB of VRAM using a batch size of 1 on the COCO validation set. We measure the average time to perform inference 
    (FPS Infer) and including post-processing (FPS).}
    \scalebox{0.7}{
    \begin{tabular}{l|c|c} \toprule
         &  \textbf{FPS Infer} & \textbf{FPS} \\ \midrule
    PN & 17.15 $\pm$ 0.52 & 7.94 $\pm$ 0.19 \\
    & [16.13, 18.71] & [7.45, 8.35] \\ \midrule
    MBv3 & 15.72 $\pm$ 0.84 & 6.63 $\pm$ 0.52 \\
    & [13.59, 18.66] & [5.60, 8.36] \\ \midrule
    R50 & 16.36 $\pm$ 0.72 & 8.16 $\pm$ 0.38 \\
    & [14.68, 18.76] & [7.03, 9.10] \\ \bottomrule
    \end{tabular}
    }
    \label{tab:app_fps}
\end{table}

\subsection{Training With A Shared Head}

Obtaining pseudo-labels using a shared head is a two-step process. First, we must train the shared head using latent sampling for a longer period of time than we would train an individual architecture. Second, we pair the shared head with an OFA body and fine-tune it to obtain a pseudo-label. We now describe the hyperparameter settings for both steps, for Detectron2 and HPE, respectively. We apply the same hyperparameter settings to bodies from all three OFA search spaces unless stated otherwise. Assume that the hyperparameters we do not mention here are the same as they are when training individual architectures. For details regarding the hyperparameter $N$, see the subsection `Latent Sampling Strategy'. 

\subsubsection{Detectron2 Latent Sampling Head}
We set $N = 3$ and train the shared head for 250k steps, while reducing the learning rate at steps 166k and 220k. Detectron2 shared heads train on a single 32GB V100 GPU using a batch size of 8 for MobileNets and 5 for ResNets. Refer to the subsection on training individual architectures for all other hyperparameters. Between sampling bodies to calculate $\vec{\mu}(x)$ and $\vec{\sigma}(x)$ and performing inference, training a sampling shared head takes around 4-5 days. 

\subsubsection{Detectron2 Pseudo-Label Fine-Tuning}
We fine-tune MobileNets for 750 steps, adjusting the learning rate at steps 465 and 635 with a batch size of 16. For ResNets, we increase the steps to 1k, adjust the learning rate at steps 620 and 850, but reduce the batch size to 12. We use 2 V100 GPUs with 32GB of VRAM when fine-tuning, which takes around 10 minutes for MobileNets and 12 for ResNets, although panoptic segmentation inference on COCO~\cite{coco} adds 5 minutes to pseudo-label an architecture. 

\subsubsection{HPE Latent Sampling Head}

When training an HPE shared head, we can pre-compute $\vec{\mu}(x)$ and $\vec{\sigma}(x)$ for all images in the LSP training set and save them to disk as a cache. This greatly speeds up shared head training as a single epoch for LSP executes in around 30 seconds on any search space. Therefore, we set $N = 5$ and train a latent sampling shared head for 5k epochs with a batch size of 256. We adopt a cosine schedule~\cite{Loshchilov2017SGDRSG} to adjust the learning rate.

\subsubsection{HPE Pseudo-Label Fine-Tuning}
We fine-tune architectures for 10 epochs using a cosine annealing learning rate schedule. The entire process takes 10-12 minutes, depending on the body.

\begin{table*}
    \centering
    \caption{Summary of statistics for latent sampling shared head architectures across all three OFA families and CV tasks in terms of the number of architectures, task performance [\%], and resource metrics such as FLOPs [G] and parameters [M]. 
    For all metrics, we report the mean and standard deviation on one line and the range on the second line.}
    \scalebox{0.7}{
    \begin{tabular}{l|c|c|c|c|c} \toprule
    \textbf{Family} & \textbf{LSP HPE} & \textbf{Obj. Det.} & \textbf{Inst. Seg.} & \textbf{Sem. Seg.} & \textbf{Pan. Seg.} \\ \midrule \midrule
    PN (\# Archs) & 2.6k & 1.6k & 1.6k & 1.6k & 1.6k \\ \midrule
    Performance [\%] & 59.54 $\pm$ 1.06 & 23.86 $\pm$ 0.68 & 23.26 $\pm$ 0.63 & 31.39 $\pm$ 0.45 & 29.73 $\pm$ 0.58  \\
     & [55.12, 62.68] & [20.68, 25.35] & [20.22, 24.66] & [29.65, 32.58] & [27.11, 31.03] \\ \midrule
    FLOPs [G]  & 15.74 $\pm$ 0.18 & 86.85 $\pm$ 1.16 & 125.53 $\pm$ 1.16 & 63.67 $\pm$ 1.16 & 174.05 $\pm$ 1.16 \\
    & [15.20, 16.32] & [83.67, 90.57] & [122.35, 129.26] & [60.49, 67.39]  & [170.87, 177.77] \\ \midrule
    Params [M]   & 12.56 $\pm$ 0.61  & 32.41 $\pm$ 0.61 & 20.75 $\pm$ 0.61 & 20.18 $\pm$ 0.61 & 37.72 $\pm$ 0.61 \\
    & [11.13, 14.38] & [31.01, 34.18] & [19.34, 22.52] & [18.77, 21.95] & [36.31, 39.49] \\ \midrule

    \midrule
    MBv3 (\# Archs)  & 2.9k & 1.3k & 1.3k & 1.3k & 1.3k \\ \midrule
    Performance [\%] & 61.55 $\pm$ 1.01 & 24.25 $\pm$ 0.61 & 24.15 $\pm$ 0.58 & 30.89 $\pm$ 0.33 & 30.12 $\pm$ 0.50 \\
    & [57.55, 63.79] & [21.92, 25.77] & [21.92, 25.63] & [29.59, 31.87] & [28.15, 31.44] \\ \midrule
    FLOPs [G] &   14.37 $\pm$ 0.17 & 84.72 $\pm$ 1.09 & 123.40 $\pm$ 1.09 & 61.54 $\pm$ 1.09 & 171.85 $\pm$ 1.09 \\
    & [13.87, 14.91] & [81.51, 88.16] & [120.19, 126.84] & [58.33, 64.98] & [168.64, 175.29] \\ \midrule
    Params [M]  & 10.23 $\pm$ 0.85 & 36.47 $\pm$ 0.83 & 24.80 $\pm$ 0.83 & 24.23 $\pm$ 0.83 & 41.77 $\pm$ 0.83 \\
    & [8.22, 13.41] & [34.49, 39.62] & [22.83, 27.96] & [22.26, 27.38] & [39.80, 44.93] \\ \midrule

    \midrule
    R50 (\# Archs) & 3.0k & 1.4k & 1.4k & 1.4k & 1.4k \\ \midrule
    Performance [\%] & 59.86 $\pm$ 2.57 & 24.72 $\pm$ 0.63 & 24.35 $\pm$ 0.60 & 30.43 $\pm$ 0.37 & 29.77 $\pm$ 0.52  \\
    & [32.27, 62.68] & [22.50, 26.33] & [22.28, 25.85] & [29.19, 31.33] & [27.73, 31.11] \\ \midrule
    FLOPs [G]  & 21.98 $\pm$ 1.86 & 132.30 $\pm$ 10.65 & 170.99 $\pm$ 10.65 & 109.13 $\pm$ 10.65 & 219.50 $\pm$ 10.65 \\
    & [17.31, 29.95] & [106.15, 181.14] & [144.83, 219.82] & [82.97, 157.96] & [193.35, 268.34] \\ \midrule
    Params [M] & 25.91 $\pm$ 6.31 & 86.23 $\pm$ 5.81 & 74.56 $\pm$ 5.81 & 73.99 $\pm$ 5.81 & 91.53 $\pm$ 5.81 \\
    & [13.40, 49.51] & [75.66, 108.80] & [64.00, 97.13] & [63.43, 96.56] & [80.97, 114.10] \\ 

    \bottomrule
    \end{tabular}
    }
    \label{tab:app_shared_head_architecture_description}
\end{table*}

\subsection{Task Head Descriptions}
Classification architectures progressively reduce the height and width of latent tensors as data passes from input to output. At the lowest resolution, the classification head will either perform a global max pooling operation or reshape the tensor to produce a long vector that can be fed into a linear layer, producing predictions $\mathbb{R}^C$ where $C$ is the number of classes. We remove the global pooling/reshaping layer and all subsequent layers to convert an IS network to perform other CV tasks. 

We enumerate the task heads we use to train architectures in greater technical detail. First, we describe the upsampling head for HPE, before continuing with the heads for Detectron2 tasks. Table~\ref{tab:app_task_summary} summarizes the properties of each task, including the size of input images, performance metrics, datasets, and input resolutions to the task head.

\subsubsection{HPE Upsampling}

HPE takes an image as input and outputs a set of joint heatmaps where each heatmap represents the model's estimation of where a given joint resides in the image. We measure 2D HPE performance using Percentage of Correct Keypoints (PCK), specifically PCKh@0.5. A joint prediction is correct if the distance between it and the ground truth is no more than half (@0.5) of the distance between the head (h) and neck joints in the same image.

After removing the global pooling and linear layer from an IC network, 
we then append a series of deconvolution operations to the end of the network. These upsample the latent tensors to the desired resolution before a $1 \times 1$ convolution produces joint heatmaps with shape $\mathbb{R}^{J \times H_J \times W_J}$ where $J$ is the number of joints and $H_J$, $W_J$ (usually 64) are the heatmap height and width, respectively. 
Compared to the standard IC benchmark ImageNet, which usually crops input images to $224 \times 224$ or less, the cropping for HPE images is slightly larger at $256 \times 256$.

\subsubsection{Feature Pyramid Networks}

We adopt Detectron2~\cite{wu2019detectron2} to implement 
OD, IS, SS and PS. 
In this framework, tasks build a top of each other, with sequential tasks borrowing modules from earlier ones. So we can 
train one architecture on multiple tasks simultaneously and obtain distinct performance labels for each. 

The first building block for all 
Detectron2 task heads is the Feature Pyramid Network (FPN)~\cite{lin2017feature}. This module is similar to the upsampling HPE head with two exceptions. First, it produces no final output, so the $1 \times 1$ convolution for that  joint heatmaps is removed. Second, the FPN connects to the base CNN architecture in several locations via multi-resolution skip-connections. Rather than only receiving the final latent representation of the 
body with the smallest resolution, FPN receives a set of inputs with distinct resolutions from the intermediate depths of the body. 
When we downsample an image or intermediate latent representation, we halve the height and width and usually increase the number of channels (the amount by which is search space-dependent). 
Specifically, let $P^\mathcal{\ell}$ denote a latent representation that we downsample $\mathcal{\ell}$ times, e.g., $P^0$ is the input image size. If this is fed into convolution with a stride of 2, the output will be $P^1$, which has half the height and width. 
FPN receives a set of input tensors $\{P^2, P^3, P^4, P^5\}$ from the body and produces an output set of tensors $\{C^2, C^3, C^4, C^5, C^6\}$. FPN computes $C^6$ by downsampling $P^5$ and applying a $1 \times 1$ convolution. 
For all resolution levels, $C^j = f^{j}(P^j) + U^{j+1}(C^{j+1})$, where $f^j$ is a $1 \times 1$ convolution and $U^{j+1}$ is an upsampling operation; we use deconvolutions in our implementation. Finally, depending on the 
search space, different $P^j$ will have a different number of channels (see Table~\ref{tab:app_family_channels}), whereas the FPN operations enforce a constant number of channels (e.g., $256$) for every $C^j$.

\subsubsection{Object Detection and Instance Segmentation}

OD and IS aim to detect a variable number of class instances (e.g., person, car, even hairdryer) in an image and either draw a bounding box or pixel mask for each instance, respectively. 
We measure OD and IS performance in terms of mean Average Precision (AP); the ability of a network architecture to correct detect the number of bounding boxes/masks as well as draw and classify them. 

We perform OD using a multi-resolution Faster R-CNN~\cite{ren2015faster}, 
which consists of a parameterized Region Proposal Network (RPN), Region of Interest (RoI) pooler, and bounding box estimator. The RPN acts as an attention mechanism, dividing the feature tensors into small grids and determining whether each unit in these grids are part of a bounding box or not. 
The RoI modules pool the attention results for a given location in an image across different resolutions and passes this information onto the estimator, which predicts the coordinates and class of each bounding box.

Mask R-CNN~\cite{he2017mask} performs IS. It builds atop the Faster R-CNN framework, specifically the RPN, using a different set of 
RoI poolers and deconvolutions 
to predict pixel masks instead of bounding boxes. 

\subsubsection{Semantic and Panoptic Segmentation}

The goal of SS~\cite{deeplabv3plus2018} is to classify every pixel in an image. We quantify SS performance using mean Intersection over Union (mIoU). Unlike OD and IS, the number of labels is known from the number of classes and size of the input image.

The SS task head is similar to that of HPE. Receiving the FPN features as input, it upsamples low-resolution features and predicts class mask predictions. We use deconvolution operations to upsample feature maps $\{C^3, C^4, C^5, C^6\}$ to the same size as $C^2$ and then use a weightless bilinear interpolation to further upsample to the original input size.

Finally, \cite{kirillov2019panoptic} define PS as the ``joint task'' of IS and SS. Like SS, PS predicts masks for each pixel in an image, but uses IS predictions to differentiate individual instances of a given class in an image, e.g., different colors for adjacent vehicles. That is, while SS would assign the same mask color to each person in a crowd of people, PS will assign different colors to each person. The PS performance metric is Panoptic Quality (PQ) which is a combination of IS class AP and SS mIoU.

\subsection{AIO-P Structure}

The backbone GNN for AIO-P consists of an embedding layer that converts node features (e.g., one-hot operation category vectors) into continuous vectors of length 32. The graph is fed into a series of 6 GNN~\cite{morris2019weisfeiler} layers, and a single graph embedding is calculated by taking the mean of all node features. Finally, an MLP regressor with 4 hidden layers and a size of 32 makes predictions. 

When adding a $K$-Adapter to the backbone, we re-use the initial embedding layer to generate continuous node features. The $K$-Adapter uses the same kind of GNN layers as the backbone; just the input dimension is doubled to 64. Likewise, the input feature size to the MLP regressor also doubles, but the hidden size remains 32, and the $K$-Adapters also average node embeddings to create a single graph embedding.

If we train multiple $K$-Adapters sequentially, they do not interface with each other when learning, only with the original backbone. Freezing the backbone weights simplifies the training process by removing the order we train each $K$-Adapter from consideration. Finally, when using multiple $K$-Adapters, we perform inference on a downstream target task by taking the average of their predictions.

\subsubsection{Predictor Training}
We train the GNN backbone for 40 epochs on a 40k training partition of our NAS-Bench-101 CG data. We use an initial learning rate of $1e^{-4}$ and a batch size of 32. $K$-Adapters train on pseudo-labeled data for 100 epochs with the same learning rate.

\subsubsection{Fine-Tuning}
We fine-tune AIO-P with or without standardization on 20 random CGs\footnote{We use random seeds to ensure the set of 20 architectures is deterministic across different runs using the same seeds.} for 100 epochs with a batch size of 1. We unfreeze the backbone weights and instead freeze the weights of any $K$-Adapters.

If we substitute standardization for AdaProxy~\cite{lu2021one}, we only fine-tune the scaling weight $\alpha$ and sparsity vector $\vec{b}$. To attain competitive performance with AdaProxy, we fine-tune for 1000 epochs and set $\lambda = 1e^{-5}$.

\subsection{Latent Sampling Strategy}

To generate $\vec{\mu}(x)$ and $\vec{\sigma}(x)$ for an image $x$, we use a subset $\mathcal{S}'$ of bodies rather than the entire search space $\mathcal{S}$ as OFA search spaces contain around $10^{18}$ bodies each. Moreover, repeated inferences on many architectures are costly, placing practical limits on the size of $\mathcal{S}'$, so pure random sampling may not guarantee an unbiased distribution of architectures. Therefore, we select $\mathcal{S}'$ using a round robin strategy with length binning. The size and behavior of an OFA architecture are sensitive to length or the number of computational blocks it contains~\cite{mills2021profiling}. We bin architectures by the number of blocks they contain. For example, for MBv3, the number of blocks is in the range $[10, 20]$, and we divide this range into length bins $\mathcal{I}$ and sample $N$ architectures per bin so that $|\mathcal{S'}| = |\mathcal{I}|N$. Re-writing Equation~\ref{eq:math_mu} with this in mind, we obtain the following: 

\begin{table}[t]
    \centering
    \caption{CV Tasks in terms of dataset, input image resolutions, latent feature resolution size between the base CNN and task head, and evaluation metric. We adopt Detectron2 notation 
    to refer to latent representation resolutions.} 
    \label{tab:app_task_summary}
    \scalebox{0.73}{
    \begin{threeparttable}
    \begin{tabular}{c|c|c|c|c} \toprule
    \textbf{Task} & \textbf{Dataset} & \textbf{Image} & \textbf{Head Input} & \textbf{Eval.}  \\ 
                  &                  & \textbf{Resolution} & \textbf{Resolution} & \textbf{Metric} \\ \midrule
    IC & ImageNet & $224^2$ & $P^5$ & Acc. \\ \midrule
    HPE & MPII/LSP & $256^2$ & $P^5$ & PCK \\ \midrule
    OD &           &          & \{$P^5$, & AP \\
    IS &  MS-COCO  &   $\sim1024^2$\tnote{$\dagger$}  & $P^4$, & AP \\
    SS &  2017    &                    & $P^3$, & mIoU \\
    PS &        &                    & $P^2$\} & PQ \\ \bottomrule
    \end{tabular}
    \begin{tablenotes}\footnotesize
	    \item{$\dagger$}Varies in aspect ratio and resolution from $640$ up to $1333$.
	\end{tablenotes}
	\end{threeparttable}
    }
\end{table}

\begin{table}[t]
    \centering
    \caption{Number of channels at each resolution level across each OFA search space. 
    For PN and MBv3, one network stage does not perform downsampling, so we concatenate the output of that stage and the previous stage for $P^4$. For R50, the number of channels is not fixed. It is an adjustable, searchable parameter corresponding to channel multipliers $\{0.65, 0.8, 1.0\}$ per resolution level.}
    \scalebox{0.73}{
    \begin{tabular}{l|c|c|c|c} \toprule
    \textbf{Family} & \textbf{$P^2$} & \textbf{$P^3$} &\textbf{$P^4$} & \textbf{$P^5$} \\ \midrule
    PN & 32 & 56 & 104$+$128$=$232 & 1664 \\ \midrule
    MBv3 & 32 & 48 & 96$+$136$=$232 & 1152 \\ \midrule
         & \{168, & \{336, & \{664, & \{1328, \\
    R50  & 208, & 408, & 816, & 1640 \\
         & 256\} & 512\} & 1024\} & 2048\} \\ \bottomrule
    \end{tabular}
    }
    \label{tab:app_family_channels}
\end{table}

\begin{equation}
    \centering
    \label{eq:code_mu}
    \vec{\mu}(x) = \dfrac{1}{|\mathcal{I}|N}\sum_{i \in \mathcal{I}}\sum_{j = 0}^{N-1}f_{B_{i, j}}(x).
\end{equation}

Specifically, $\mathcal{I} = 5$ in our experiments. Therefore, the set of bins and their ranges for MBv3 and R50 are $\{[10, 11], [12, 13], [14, 16], [17, 18], [19, 20]\}$. For PN, 
the block ranges are slightly different, e.g., $[11, 21]$, 
so we shift the bin ranges accordingly. 
When training a shared head, we cycle through the bins with every 
minibatch and replace the oldest architecture in that bin. If the shared head requires multiple feature resolutions, e.g., FPN, we use the same architecture set to compute $\vec{\mu}(x)$ and $\vec{\sigma}(x)$ for each size, but sample using unique $\zeta$ at each scale. 

\subsubsection{Preliminary ImageNet Experiment}

We perform a preliminary experiment to test the efficacy of the latent sampling concept. Specifically, we perform inference on the ImageNet validation set using this technique by sampling the final latent representation before the IC head, and feeding the result into the IC head of the largest body in the search space. Table~\ref{tab:app_sample_imagenet} lists the results, showing that our latent sampling approach can achieve top-1 accuracy performance comparable to what we observe evaluating individual networks.

\subsubsection{Additional Correlation Statistics}

Table~\ref{tab:sample_srcc} shows the performance distribution and SRCC when a set of architectures are individually trained and fine-tuned using a shared head. We repeat this routine for OD/IS/SS/PS using a latent sampling shared head to calculate the SRCC between pseudo-labels and the ground truth for Detectron2 tasks. We enumerate the results in Table~\ref{tab:app_sample_srcc}.

\subsubsection{Body Swapping}
We perform body swapping by changing the body feature extractor paired with the shared head 10 batches. We leverage the pre-trained OFA weights and only update the randomly initialized head weights when training the shared head. 

\begin{table}[H]
    \centering
    \caption{Average ImageNet top-1 accuracy of a handful of randomly selected individual architectures with our latent sampling idea. We set $|\mathcal{B}| = N = 5$.}
    \label{tab:app_sample_imagenet}
    \scalebox{0.73}{
    \begin{tabular}{l|c|c|c} \toprule
         &  \textbf{PN} & \textbf{MBv3} & \textbf{R50} \\ \midrule
    Ind. Archs &  75.41\% & 76.94\% & 78.19\% \\
    Sampling   &  76.93\% & 77.68\% & 77.12\% \\ \bottomrule
    \end{tabular}
    }
\end{table}

\begin{table}[H]
    \centering
    \caption{SRCC values between individually trained architecture performance and when the same architectures are fine-tuned using a latent sampling shared head.} 
    \scalebox{0.73}{
    \begin{tabular}{l|c|c|c} \toprule
    \textbf{Task}&  \textbf{PN} & \textbf{MBv3} & \textbf{R50} \\ \midrule
    LSP-HPE & 0.659 & 0.576 & 0.375 \\ 
    Obj. Det. & 0.622 & 0.570 & 0.501 \\
    Inst. Seg. & 0.557 & 0.566 & 0.512 \\
    Sem. Seg. & 0.520 & 0.310 & 0.244 \\
    Pan. Seg. & 0.505 & 0.510 & 0.603 \\
    \bottomrule
    \end{tabular}
    }
    \label{tab:app_sample_srcc}
\end{table}

\subsubsection{ResNet50 Shared Head}

The main difference between a shared head and the head of an individually trained architecture is how we train them.
Structurally, they are identical, at least for PN and MBv3. However, this is not the case for R50, as the number of channels in a latent representation changes depending on the body's structure~\cite{mills2021profiling}. Table~\ref{tab:app_family_channels} lists the channel sizes for all three families. Because the number of channels in R50 is not fixed as it is for PN and MBv3, when designing a shared head for R50, we set the number of channels per resolution level to be the maximum possible, zero-pad tensors with fewer channels and then apply a 1 $\times$ 1 convolution prior to FPN processing.

\begin{table*}[ht!]
    \centering
    \caption{MAE [\%] of AIO-P on three search spaces and six tasks in the fine-tuning setting, compared to 
    GNN without and with rescaling by Eqs.~\ref{eq:standardization}, \ref{eq:accTrans}. AIO-P adopts 2 $K$-Adapters, trained on LSP and OD. AIO-P uses Equation~\ref{eq:accTrans} and standardize regression targets. Results averaged across 5 seeds. }
    \scalebox{0.73}{
    \begin{tabular}{l|ccc|ccc|ccc} \toprule
    &  & \textbf{ProxylessNAS} &  &  & \textbf{MobileNetV3} & & & \textbf{ResNet-50} &  \\ \midrule
    \textbf{Task} & GNN & +Eqs.~\ref{eq:standardization} \& \ref{eq:accTrans} & AIO-P &  GNN & +Eqs.~\ref{eq:standardization} \& \ref{eq:accTrans} & AIO-P &  GNN & +Eqs.~\ref{eq:standardization} \& \ref{eq:accTrans} & AIO-P \\ \midrule
    LSP  & 0.55 $\pm$ 0.39\% & 0.56 $\pm$ 0.04\%& \textbf{0.48}$\pm$ 0.02\% & 0.70 $\pm$ 0.14\% & 0.57$\pm$ 0.01\%& \textbf{0.52} $\pm$ 0.01\% & \textbf{0.81} $\pm$ 0.11\% & 0.93$\pm$ 0.05\% & 0.93 $\pm$ 0.05\% \\
    MPII & 0.43 $\pm$ 0.22\% & 0.28 $\pm$ 0.02\%& \textbf{0.26}$\pm$ 0.02\% & 0.33 $\pm$ 0.03\% & 0.28$\pm$ 0.02\%& \textbf{0.26} $\pm$ 0.01\% & \textbf{0.28} $\pm$ 0.03\% & 1.11$\pm$ 0.53\% & 1.02 $\pm$ 0.07\% \\
    OD   & 0.90 $\pm$ 0.16\% & 0.74 $\pm$ 0.07\%& \textbf{0.53}$\pm$ 0.04\% & 0.69 $\pm$ 0.13\% & 0.78$\pm$ 0.05\%& \textbf{0.56} $\pm$ 0.08\% & 0.89 $\pm$ 0.19\% & 0.64$\pm$ 0.05\% & \textbf{0.50} $\pm$ 0.06\% \\
    IS   & 0.72 $\pm$ 0.15\% & 0.75 $\pm$ 0.09\%& \textbf{0.33}$\pm$ 0.03\% & 0.66 $\pm$ 0.25\% & 0.56$\pm$ 0.04\%& \textbf{0.40} $\pm$ 0.02\% & 0.61 $\pm$ 0.08\% & 0.54$\pm$ 0.03\% & \textbf{0.41} $\pm$ 0.05\% \\ 
    SS   & 0.68 $\pm$ 0.12\% & 0.58 $\pm$ 0.04\%& \textbf{0.33}$\pm$ 0.03\% & 0.93 $\pm$ 0.19\% & 0.61$\pm$ 0.06\%& \textbf{0.43} $\pm$ 0.03\% & 0.65 $\pm$ 0.21\% & 0.47$\pm$ 0.01\% & \textbf{0.43} $\pm$ 0.02\% \\
    PS   & 0.53 $\pm$ 1.00\% & 0.62 $\pm$ 0.04\%& \textbf{0.33}$\pm$ 0.04\% & 0.64 $\pm$ 0.17\% & 0.61$\pm$ 0.03\%& \textbf{0.43} $\pm$ 0.03\% & 0.71 $\pm$ 0.19\% & 0.43$\pm$ 0.04\% & \textbf{0.38} $\pm$ 0.04\% \\ \bottomrule
    \end{tabular}
    }
    \label{tab:app_main_mae}
\end{table*}

\begin{table*}[ht!]
    \centering
    \caption{SRCC on AIO-P on three search spaces and size tasks. 
    Same configurations as Table~\ref{tab:app_main_mae}.}
    \scalebox{0.73}{
    \begin{tabular}{l|ccc|ccc|ccc} \toprule
    &  & \textbf{ProxylessNAS} &  &  & \textbf{MobileNetV3} & & & \textbf{ResNet-50} &  \\ \midrule
    \textbf{Task} & GNN & +Eqs.~\ref{eq:standardization} \& \ref{eq:accTrans} & AIO-P &  GNN & +Eqs.~\ref{eq:standardization} \& \ref{eq:accTrans} & AIO-P &  GNN & +Eqs.~\ref{eq:standardization} \& \ref{eq:accTrans} & AIO-P \\ \midrule
    LSP  & 0.610 $\pm$ 0.018 & 0.583$\pm$ 0.068& \textbf{0.668}$\pm$ 0.034 & 0.449 $\pm$ 0.057 & 0.435$\pm$ 0.090& \textbf{0.567}$\pm$ 0.014 & \textbf{0.397} $\pm$ 0.196 & 0.314$\pm$ 0.059& 0.265$\pm$ 0.021 \\
    MPII & 0.770 $\pm$ 0.017 & \textbf{0.803}$\pm$ 0.015& 0.773$\pm$ 0.019 & 0.680 $\pm$ 0.052 & 0.732$\pm$ 0.099& \textbf{0.744}$\pm$ 0.053 & \textbf{0.700} $\pm$ 0.072 & 0.531$\pm$ 0.071& 0.535$\pm$ 0.023 \\
    OD   & 0.304 $\pm$ 0.460 & 0.589$\pm$ 0.059& \textbf{0.800}$\pm$ 0.048 & 0.283 $\pm$ 0.395 & 0.374$\pm$ 0.126& \textbf{0.703}$\pm$ 0.058 & 0.526 $\pm$ 0.040 & 0.668$\pm$ 0.046& \textbf{0.871}$\pm$ 0.022 \\ 
    IS   & 0.277 $\pm$ 0.696 & 0.330$\pm$ 0.140& \textbf{0.894}$\pm$ 0.033 & 0.547 $\pm$ 0.151 & 0.505$\pm$ 0.083& \textbf{0.791}$\pm$ 0.028 & 0.611 $\pm$ 0.026 & 0.590$\pm$ 0.031& \textbf{0.881}$\pm$ 0.024 \\ 
    SS   & 0.195 $\pm$ 0.328 & 0.562$\pm$ 0.108& \textbf{0.849}$\pm$ 0.033 & 0.447 $\pm$ 0.140 & 0.653$\pm$ 0.061& \textbf{0.822}$\pm$ 0.028 & 0.552 $\pm$ 0.058 & 0.653$\pm$ 0.023& \textbf{0.677}$\pm$ 0.032 \\ 
    PS   & 0.741 $\pm$ 0.043 & 0.297$\pm$ 0.083& \textbf{0.868}$\pm$ 0.036 & 0.568 $\pm$ 0.147 & 0.373$\pm$ 0.032& \textbf{0.786}$\pm$ 0.046 & 0.601 $\pm$ 0.092 & 0.696$\pm$ 0.067& \textbf{0.858}$\pm$ 0.031 \\ \bottomrule
    \end{tabular}
    }
    \label{tab:app_main_srcc}
\end{table*}

\subsection{Computational Graph Structure}
In a CG, each node is an irreducible primitive operation, e.g., tensor operations like mean and concat, weighted layers like convolutions and linear layers, activation functions like ReLU or even upsampling operations. This is in contrast to the pre-defined operation `bundles', e.g., the `Nor\_Conv' operations of NAS-Bench-201~\cite{dong2020nasbench201}, itself a proxy for the sequence `ReLU-Convolution-BatchNorm' which a CG would represent as a subgraph of 3 nodes and 2 directed edges. Node features include the height-width and channel dimensions of the input and output latent tensors, a feature for weight tensor dimensions if applicable (e.g., for convolutions, but not for ReLU), a Boolean on whether bias is enabled, and a one-hot category for the operation type. 

\subsection{Additional Results}

Tables~\ref{tab:app_main_mae} and \ref{tab:app_main_srcc} compare AIO-P to the backbone GNN with and without Equations~\ref{eq:standardization} and \ref{eq:accTrans}, this time in the fine-tuning context (whereas before, Tables~\ref{tab:main_mae} and \ref{tab:main_srcc} only consider the zero-shot transfer). AIO-P now achieves the best MAE performance on the PN and MBv3 search spaces and the best SRCC on MBv3. While the GNN can attain high metrics for LSP and MPII in the R50 search space, these are limited results. It cannot generalize that performance to other tasks and search spaces, and the tasks it does well on are also tasks with high PCK ranges similar to classification accuracy, as Table~\ref{tab:app_full_architecture_description} shows.

\subsection{Hardware and Software Details}

Our experimental servers have 8 NVIDIA Tesla V100 GPUs with 32GB of VRAM per card coupled with an Intel Xeon Gold 6140 GPU and 756GB of RAM. Each server runs Ubuntu 20.04.4 LTS. 
We create Compute Graphs using models from \texttt{TensorFlow==1.15.0}, specifically, replicating \texttt{PyTorch} models with the \texttt{Keras} API if necessary. 

We train architectures using \texttt{PyTorch==1.8.1} and \texttt{Detectron2==0.6} where applicable and use \texttt{PyTorch-Geometric} to train GNNs.

\end{document}